\documentclass[11pt]{article}

\usepackage[preprint]{acl}

\usepackage{times}
\usepackage{latexsym}

\usepackage[T1]{fontenc}

\usepackage[utf8]{inputenc}

\usepackage[whole]{bxcjkjatype}

\usepackage{microtype}
\usepackage{ulem}
\usepackage{tipa}

\usepackage{inconsolata}

\usepackage{graphicx}
\usepackage{subcaption}
\usepackage{enumitem}
\usepackage{pifont}
\usepackage[most]{tcolorbox}
\usepackage{longtable}
\usepackage{array}
\usepackage{xltabular}
\usepackage{refcount}

\usepackage{amsmath}
\usepackage{amsthm}
\usepackage{booktabs}
\usepackage{multirow}
\usepackage{caption}
\usepackage{amssymb}
\usepackage[table,xcdraw,dvipsnames]{xcolor}

\newcommand{\citedblock}[1]{%
\vspace{0.2em}
\noindent
{\color{black!40}\vrule width 1.5pt}\hspace{0.8em}%
\parbox[t]{\dimexpr\linewidth-2pt-0.8em\relax}{#1}
}

\title{Hallu\underline{Citation} Matters: Revealing the Impact of Hallucinated References with 300 Hallucinated Papers in ACL Conferences}

\author{Yusuke Sakai, \;\;
        Hidetaka Kamigaito, \;\;
        Taro Watanabe \\
  Nara Institute of Science and Technology (NAIST), Japan  \\
  \texttt{\{sakai.yusuke.sr9, kamigaito.h, taro\}@is.naist.jp}}

\begin{document}
\maketitle %
\begin{abstract}

Recently, we have often observed hallucinated citations or references that do not correspond to any existing work in papers under review, preprints, or published papers. Such hallucinated citations pose a serious concern to scientific reliability. When they appear in accepted papers, they may also negatively affect the credibility of conferences. In this study, we refer to hallucinated citations as \textit{\textbf{``Hallu\underline{Citation}''}} and systematically investigate their prevalence and impact.
We analyze all papers published at ACL, NAACL, and EMNLP in 2024 and 2025, including main conference, Findings, and workshop papers. Our analysis reveals that nearly 300 papers contain at least one HalluCitation, most of which were published in 2025. Notably, half of these papers were identified at EMNLP 2025, the most recent conference, indicating that this issue is rapidly increasing. Moreover, more than 100 such papers were accepted as main conference and Findings papers at EMNLP 2025, affecting the credibility. %

\end{abstract}

\section{Introduction}

\begin{figure}[!t]
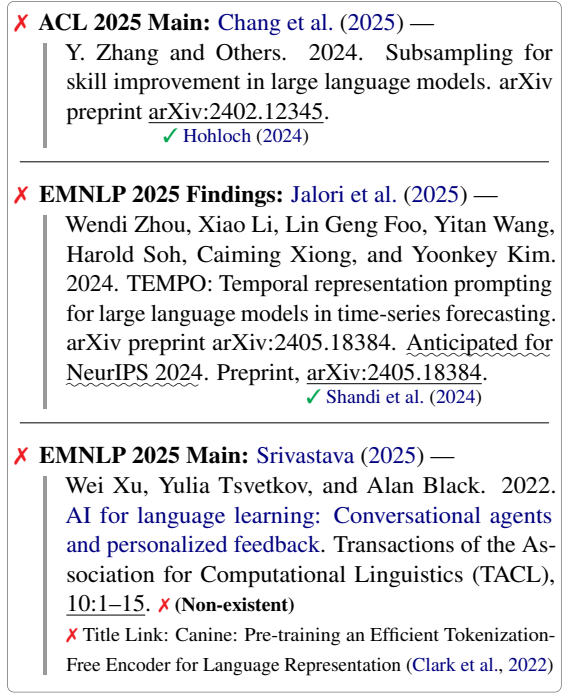

    \centering
    \resizebox{0.915\linewidth}{!}{%
\begin{tcolorbox}[
  width=1.1\linewidth,
  colback=white,
  colframe=black!60,
  boxrule=0.3pt,   %
  left=0.5em, right=0.5em, boxsep=0pt,
  top=0.5em, bottom=0.5em,
  grow to left by=2mm,
  grow to right by=2mm,
]
\begin{itemize}[leftmargin=*,  itemsep=0.3em, topsep=0.5em, left=0em]
  \item[\textcolor{Red}{\ding{55}}] \textbf{ACL 2025 Main:} \citet{chang-etal-2025-automixer} ---\\
  \citedblock{Y. Zhang and Others. 2024. Subsampling for skill improvement in large language models. arXiv preprint $\underset{\text{\small\textcolor{Green}{\ding{51}} \citet{hohloch2025homoclinicfloerhomologydirect}}}{\underline{\text{arXiv:2402.12345}}}$.}
  \par\smallskip\hrule\smallskip
  \item[\textcolor{Red}{\ding{55}}] \textbf{EMNLP 2025 Findings:} \citet{jalori-etal-2025-flairr} ---\\
  \citedblock{Wendi Zhou, Xiao Li, Lin Geng Foo, Yitan Wang, Harold Soh, Caiming Xiong, and Yoonkey Kim. 2024. TEMPO: Temporal representation prompting for large language models in time-series forecasting. arXiv preprint arXiv:2405.18384. \uwave{Anticipated for NeurIPS 2024}. Preprint, $\underset{\text{\small\textcolor{Green}{\ding{51}} \citet{shandi2024decentralized}}}{\underline{\text{arXiv:2405.18384}}}$.}
  \par\smallskip\hrule\smallskip
  \item[\textcolor{Red}{\ding{55}}] \textbf{EMNLP 2025 Main:} \citet{srivastava-2025-large} ---\\
  \citedblock{Wei Xu, Yulia Tsvetkov, and Alan Black. 2022. \href{https://www.aclweb.org/anthology/2022.tacl-1.5}{AI for language learning: Conversational agents and personalized feedback}. Transactions of the Association for Computational Linguistics (TACL), \underline{10:1–15}. {\small\textcolor{Red}{\ding{55}} \textbf{(Non-existent)}}\\ {\small \textcolor{Red}{\ding{55}}  Title Link: Canine: Pre-training an Efficient Tokenization-Free Encoder for Language Representation \cite{clark-etal-2022-canine}}}
\end{itemize}

\end{tcolorbox}
}%
    \caption{Examples of incorrect reference information. Some references include incorrect arXiv IDs or contain uncertain phrases such as ``Anticipated for''. Even references that appear plausible may contain HalluCitations, e.g., incorrect link information or references to pages where the cited paper does not exist, as in the last case.}
    \label{fig:examples}
\end{figure}

Thanks to the advances in AI for Science research~\cite{scieval_2024, chen2025ai4researchsurveyartificialintelligence, shi2025deepresearchsystematicsurvey} and AI Scientist technologies~\cite{wei2025aiscienceagenticscience, huang2025deepresearchagentssystematic, yamada2025aiscientistv2workshoplevelautomated, tang2025airesearcherautonomousscientificinnovation, shao2025omniscientistcoevolvingecosystemhuman, xu2025probingscientificgeneralintelligence}, academic research and writing have shifted from traditionally human-centered processes to ones supported by and actively leveraging AI systems such as large language models (LLMs).
In particular, LLMs are enabling a wide range of tools for writing assistance~\cite{10.1145/3711020}, literature search~\cite{asai2024openscholarsynthesizingscientificliterature}, citation recommendation~\cite{sahinuc-etal-2024-systematic, celik-tekir-2025-citebart}, idea discussion~\cite{zheng-etal-2025-deepresearcher}, peer-review support and analysis~\cite{jin-etal-2024-agentreview, pmlr-v267-starace25a, hossain-etal-2025-llms}, and other scientific automation~\cite{sahinuc-etal-2024-efficient}.
These benefits have been especially significant for non-native English speakers, as AI tools can improve translation and grammatical quality and help create fairer opportunities for participation in the international academic community~\cite{Lepp_2025}.

However, while these advances have boosted research productivity, the research community has begun to face new challenges arising from the exponential growth in paper submissions, particularly in securing qualified reviewers. As a result, the qualification bar has been lowered\footnote{\label{footnote:review}\url{https://aclrollingreview.org/incentives2025}}, and reviewers are often required to evaluate many papers on unfamiliar topics within a limited time. Moreover, recent review processes involve potential penalties for delayed reviews, such as desk rejection of a reviewer's own submissions\footref{footnote:review}. Under such pressure and concern, the focus of peer review has shifted from conducting thorough and rigorous evaluations to completing reviews under strict time constraints.
Consequently, we face growing difficulty in carefully inspecting the rigor and factual correctness of submitted manuscripts.
In particular, as shown in Figure~\ref{fig:examples}, we have recently observed non-existent citations, i.e., hallucinated references that do not correspond to any real prior work. Such false citations pose a serious threat to scientific reliability and undermine respect for existing work. When they appear in published papers, they may also negatively affect the overall quality and credibility of academic conferences and journals.

In this study, we refer to such hallucinated references as \textit{\textbf{``Hallu\underline{Citation}''}} (Hallucination + Citation, \textipa{/h@ˌlu:.saIˈteI.S@n/}). Papers that include at least one HalluCitation are referred as \textit{\textbf{``HalluCited''}} papers (Hallucination + Cited, \textipa{/ˈh\ae l.u:ˌsaI.tId/}). We systematically investigate their prevalence and impact at top-tier NLP conferences.
We analyze over 17,000 papers published at ACL, NAACL, and EMNLP in 2024 and 2025, including main conference, Findings, and workshop papers. Our analysis and discussion reveal the following findings:

\begin{itemize}[leftmargin=*, itemsep=0em, topsep=0.5em, left=0em]
\item Almost 300 papers contain at least one HalluCitation, mostly published in 2025. Notably, half of these papers were identified at EMNLP 2025, the most recent conference, indicating that this issue is rapidly increasing. Moreover, over 100 such papers are accepted as main and Findings papers in EMNLP 2025, affecting the credibility.
\item We propose a HalluCitation detection method based on OCR and database matching, and show that it effectively supports our analysis in identifying actual HalluCitations. In particular, papers with four or more detected candidates exhibit a high incidence of HalluCitations, suggesting a practical guideline for integration into automated detection toolkits.
\item Even when HalluCitations are present, they cannot be immediately attributed to AI-generated content, as they may also originate from secondary sources, making their causes more complex. Accordingly, the presence of HalluCitations alone should not lead to immediate penalties for authors. Instead, it is important to establish prior consensus on preventive measures, such as author toolkits for pre-submission checks.
\end{itemize}

\section{Analysis Methods}

\begin{figure}
    \centering
    \includegraphics[width=\linewidth]{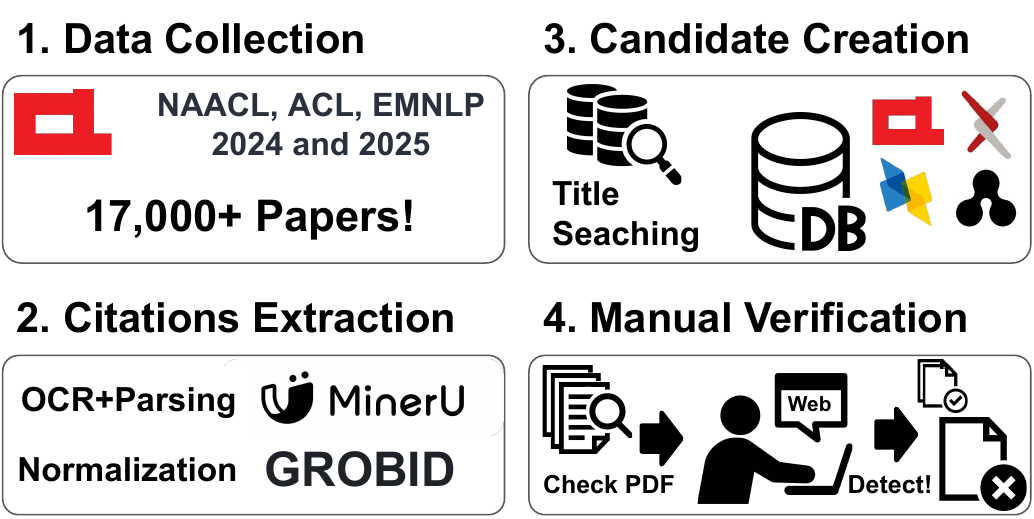}
    \caption{Our analysis method. First, we extract citations from the collected PDFs. Next, we attempt to identify the corresponding papers in a reference paper database using fuzzy matching on citation titles, and flag citations for which no matching paper can be found as HalluCitation candidates. Finally, we manually verify the existence of each candidate by referring to the original PDFs and conducting web searches. If it cannot be objectively identified, we mark it as a HalluCitation.}
    \label{fig:proceess}
\end{figure}

This investigation is similar to finding a needle in a haystack, as it requires checking an enormous number of citations to detect a small number of HalluCitation. Therefore, manually verifying all citations is impractical. To address this challenge, as shown in Figure~\ref{fig:proceess}, we exploit the assumption that \textit{``the majority of citations are correct''} and systematically list up candidate HalluCitation. Then, we manually verify each candidate to identify HalluCitation.
Note that, since the actual number of HalluCitation is unknown, our reported results should be interpreted as a lower bound, and additional HalluCitation may exist in the wild.

\subsection{Data Collection}

\begin{table}[t]
    \centering
    \small
    \setlength{\tabcolsep}{3.5pt}
    \rowcolors{2}{gray!10}{white}
    \begin{tabular}{ccrrrrr} \toprule
         Year & Venue & Main & Findings & S.I.D. & WS & Total \\ \midrule
        2024 & NAACL & 562 & 296 & 94 & 609 & 1,561 \\
        2024 &   ACL & 941 & 975 & 84 & 721 & 2,721 \\
        2024 & EMNLP & 1,268 & 1,003 & 173 & 510 & 2,954 \\
        2025 & NAACL & 718 & 474 & 177 & 560 & 1,929 \\
        2025 &   ACL & 1,699 & 1,387 & 259 & 1,165 & 4,510 \\
        2025 & EMNLP & 1,809 & 1,405 & 269 & 684 & 4,167 \\ \midrule
        \multicolumn{2}{c}{Overall} & 6,997 & 5,540 & 1,056 & 4,249 & 17,842\\
        \bottomrule
    \end{tabular}
    \caption{Number of PDFs. S.I.D. denotes the total number of papers from the Student Research Workshop, Industry Track, and Demonstration Track. Note that the Industry Track was not held at ACL 2024, and the Student Research Workshop was not held at EMNLP. Conferences are ordered chronologically by their dates.}
    \label{tab:data_stats_pdf}
\end{table}

In this study, we target all papers presented at a total of six conferences, NAACL, ACL, and EMNLP, held in 2024 and 2025, including main conference papers, Findings papers, and all workshop papers. We collected all PDF files and metadata of archival papers registered as events in the ACL Anthology\footnote{\url{https://aclanthology.org/}}, excluding tutorials and proceedings. Table~\ref{tab:data_stats_pdf} presents the statistics of the collected PDFs. In total, our dataset covers over 17,000 papers.

\subsection{Citation Extraction}

\begin{table}[t]
    \centering
    \small
    \setlength{\tabcolsep}{2.5pt}
    \rowcolors{2}{gray!10}{white}
    \begin{tabular}{ccrrrrrr} \toprule
        Year & Venue &  Mean &  Std. & Q1 & Q2 & Q3 & Total Citations \\ \midrule
        2024 & NAACL & 36.33 & 20.48 & 21 & 35 & 48 &  56,717  \\
        2024 &   ACL & 41.26 & 24.04 & 27 & 39 & 52 & 112,262  \\
        2024 & EMNLP & 42.91 & 23.90 & 29 & 40 & 53 & 126,744 \\
        2025 & NAACL & 39.33 & 22.42 & 25 & 37 & 50 &  75,870 \\
        2025 &   ACL & 42.63 & 25.94 & 27 & 40 & 54 & 192,245 \\
        2025 & EMNLP & 42.67 & 27.68 & 28 & 39 & 52 & 177,818  \\ \midrule
        \multicolumn{2}{c}{Overall} & 41.57 & 25.03 & 27 & 39 & 52 & 741,656 \\
        \bottomrule
    \end{tabular}
    \caption{Number of extracted citations. We report the mean, standard deviation, quartiles (Q1, Q2=median, Q3), and the total number of citations for each venue.}
    \label{tab:data_stats_citation}
\end{table}

Next, we extract citation information from the PDFs.
Through pilot studies, we found that a pipeline combining OCR-based extraction of raw bibliographical reference strings with normalization of the corresponding bibliographical references yields the most efficient citation extraction.
Accordingly, first, we use MinerU~\cite{mineru} to extract reference entries from the references section at the text-block level. Then, we apply GROBID~\cite{GROBID} to parse and normalize each reference entry, obtaining structured bibliographic information such as titles\footnote{Although GROBID can directly extract bibliographic information from PDFs, our pilot studies indicate that its citation extraction is not sufficiently robust for ACL-style reference formats. Specifically, when reference lists span multiple pages, GROBID frequently fails to extract references appearing on subsequent pages, leading to under-coverage. For instance, in some cases, only around 10 citations are extracted from papers that actually contain 50 references. To mitigate this, we explicitly provide reference blocks as input to GROBID.}.
Table~\ref{tab:data_stats_citation} shows the statistics of the extracted citation counts. From these results, we observe that the median number of citations per paper is 39 and that the quartiles remain stable across venues. In total, we extracted over 740,000 citations.

\subsection{Canditate Creation}
\label{sec:candidate-creation}

\begin{table}[t]
    \centering
    \small
    \setlength{\tabcolsep}{2.5pt}
    \rowcolors{2}{gray!10}{white}
    \begin{tabular}{ccrrrr} \toprule
        Year & Venue & \multicolumn{1}{c}{\#Papers} & \multicolumn{1}{c}{\#Citations} &   Ave. & Max  \\ \midrule
        2024 & NAACL &                211 (13.52\%) &     268 (0.47\%) &   1.27 &   6  \\
        2024 &   ACL &                381 (14.00\%) &     476 (0.42\%) &   1.25 &   4  \\
        2024 & EMNLP &                443 (15.00\%) &     556 (0.44\%) &   1.26 &   7  \\
        2025 & NAACL &                336 (17.42\%) &     465 (0.61\%) &   1.38 &  10  \\
        2025 &   ACL &                796 (17.65\%) &   1,117 (0.58\%) &   1.40 &  17  \\
        2025 & EMNLP &                783 (18.79\%) &   1,222 (0.69\%) &   1.56 &  31  \\ \midrule
        \multicolumn{2}{c}{2024} &  1,035 (14.30\%) &   1,300 (0.44\%) &   1.26 &   7  \\ 
        \multicolumn{2}{c}{2025} &  1,915 (18.06\%) &   2,804 (0.63\%) &   1.46 &  31  \\ \midrule
     \multicolumn{2}{c}{Overall} &  2,950 (16.53\%) &   4,104 (0.55\%) &   1.39 &  31  \\
        \bottomrule
    \end{tabular}
    \caption{Number of candidate papers and citations. We also report the proportion of candidate papers and citations relative to the total. Ave. denotes the average number of candidate citations per paper, while Max denotes the maximum number of candidate citations observed in a single paper.}
    \label{tab:data_stats_candidate}
\end{table}

As shown in Table~\ref{tab:data_stats_citation}, the total number of extracted citations is extremely large, making it impractical to manually inspect all citations. Therefore, we focus on citations that explicitly indicate references to papers from the ACL Anthology or arXiv, and extract a subset of candidate citations by matching them against reference databases.

First, we heuristically narrow down the target citations by checking whether the raw citation strings contain keywords related to ACL conferences or arXiv, such as ACL, EMNLP, or arXiv.
Next, we search for corresponding papers primarily using metadata from the ACL Anthology\footnote{\url{https://github.com/acl-org/acl-anthology/tree/c9d3481/data}} and arXiv\footnote{\url{https://www.kaggle.com/datasets/Cornell-University/arxiv/versions/263}}.
In addition, since this heuristic filtering serves as a coarse screening, citations to papers outside ACL venues or arXiv may still remain. To further filter such cases, we additionally leverage DBLP\footnote{\url{https://drops.dagstuhl.de/entities/artifact/10.4230/dblp.xml.2025-12-01}} and OpenAlex~\cite{openalex}\footnote{We use the REST APIs via \texttt{PyAlex}~\cite{pyalex}.}.

For database matching, we employ character-level fuzzy matching on citation titles using a similarity score based on the normalized Levenshtein distance with the \texttt{RapidFuzz} library~\cite{rapidfuzz}. Specifically, we compute title similarity as $1 - d_{\text{lev}} / \max(|s_1|, |s_2|)$, where $s_1$ and $s_2$ are the character strings being compared, $d_{\text{lev}}$ denotes the Levenshtein edit distance~\cite{levenshtein1966binary}, and scale it to the range $[0, 1]$.
If no title in the reference databases achieves a similarity score of $0.9$ or higher, we regard the citation as potentially non-existent and include it in a list of candidates for manual verification.

\paragraph{Observation.}
Table~\ref{tab:data_stats_candidate} shows the statistics of candidate citations and papers. From these results, we observe that both the average number of candidate citations per paper and the maximum number increase over time. In addition to the raw counts of citations and papers, the corresponding percentages also show a consistent upward trend. In particular, when comparing 2024 and 2025, we observe different trends in the proportions of targeted papers and citations.
These observations suggest that, although our coarse screening and simple database matching may introduce some noise, the increase in candidate citations exceeds what can be explained by such methodological noise alone. This indicates that factors beyond the limitations of our method are likely contributing to the observed increase, most notably the effect of \textbf{\textit{HalluCitation}}.

\subsection{Manual Verification}

Finally, we manually verify all candidate citations identified in Section~\ref{sec:candidate-creation} to determine whether each citation refers to an actual existing work. Since our goal is to detect papers that contain HalluCitation, we mark a paper as \textit{``HalluCited''} immediately when at least one HalluCitation is identified and terminate further checks for that paper. We apply this procedure to all 2,950 target papers.

If identifiable information, such as a clickable link, DOI, arXiv ID, page numbers, or venue information, is provided in the citation, we use this information to search for the referenced paper. If such information is missing or insufficient to locate the paper, we conduct a web search using the citation title to find potentially corresponding works.
We classify a citation as ``HalluCitation'' if no corresponding paper can be found, e.g., when a search returns only the target paper itself, or if a similar paper is found but at least two key attributes, such as the title, authors, venue, or page numbers, do not match any reliable source. This verification procedure adopts a conservative criterion that basically trusts the papers, prioritizing precision over recall.
Appendix~\ref{sec:list-of-papers} shows the list of HalluCited papers.

\section{Results}

\begin{table}[t]
    \centering
    \small
    \setlength{\tabcolsep}{2pt}
    {\renewcommand{\arraystretch}{1.5}
    \setlength{\aboverulesep}{0pt}
    \setlength{\belowrulesep}{0pt}
    \rowcolors{2}{gray!10}{white}
    \begin{tabular}{ccrrrrr} \toprule
         Year & Venue &  \multicolumn{1}{c}{Main} &    \multicolumn{1}{c}{Findings} &      \multicolumn{1}{c}{S.I.D.} &          \multicolumn{1}{c}{WS} &        \multicolumn{1}{c}{Total} \\ \midrule
         2024 & NAACL &  $1\ ^{0.18}_{50.0}$ &  $0\ ^{0.00}_{0.00}$ &  $0\ ^{0.00}_{0.00}$ &  $1\ ^{0.16}_{50.0}$ &   $2\ ^{0.12}_{-}$ \\ 
         2024 &   ACL &  $3\ ^{0.31}_{42.9}$ &  $1\ ^{0.10}_{14.3}$ &  $0\ ^{0.00}_{0.00}$ &  $3\ ^{0.42}_{42.9}$ &   $7\ ^{0.26}_{-}$ \\ 
         2024 & EMNLP &  $3\ ^{0.24}_{27.3}$ &  $4\ ^{0.40}_{36.4}$ &  $2\ ^{1.16}_{18.2}$ &  $2\ ^{0.39}_{18.2}$ &  $11\ ^{0.37}_{-}$ \\ \midrule
         2025 & NAACL &  $6\ ^{0.84}_{17.1}$ & $10\ ^{2.11}_{28.6}$ &  $2\ ^{1.13}_{5.71}$ & $17\ ^{3.04}_{48.6}$ &  $35\ ^{1.81}_{-}$ \\
         2025 &   ACL & $14\ ^{0.82}_{16.3}$ & $18\ ^{1.30}_{21.0}$ & $10\ ^{3.86}_{11.6}$ & $44\ ^{3.78}_{51.2}$ &  $86\ ^{1.91}_{-}$ \\
         2025 & EMNLP & $60\ ^{3.32}_{39.0}$ & $45\ ^{3.20}_{29.2}$ & $13\ ^{4.83}_{8.44}$ & $36\ ^{5.26}_{23.4}$ & $154\ ^{3.70}_{\ -}$ \\ \midrule
    \multicolumn{2}{c}{2024} &  $7\ ^{0.25}_{35.0}$ &  $5\ ^{0.22}_{25.0}$ &  $2\ ^{0.57}_{10.0}$ &  $6\ ^{0.33}_{30.0}$ &  $20\ ^{0.28}_{-}$ \\
    \multicolumn{2}{c}{2025} & $80\ ^{1.89}_{29.1}$ & $73\ ^{2.24}_{26.5}$ & $25\ ^{3.55}_{9.09}$ & $97\ ^{4.03}_{35.3}$ &  $275\ ^{2.59}_{-}$ \\ \midrule
    \multicolumn{2}{c}{Overall} & $87\ ^{1.24}_{29.5}$ & $78\ ^{1.41}_{26.4}$ & $27\ ^{2.56}_{9.15}$ &  $103\ ^{2.42}_{34.9}$ &  $295\ ^{1.65}_{-}$ \\
        \bottomrule 
    \end{tabular}
    }
    \caption{Number of papers including \textbf{\textit{``HalluCitation''}}. We also report the proportion of papers with HalluCitation relative to the total number of papers, as well as track-specific proportions within each venue. We denote these values as $\mathrm{NUM}^{\text{Venue}\%}_{\text{Track}\%}$. Note that in some venues and tracks with only a small number of papers, the percentages may be sensitive to individual papers. Nevertheless, they remain useful for tracking the trends.}
    \label{tab:data_stats_HalluCitations}
\end{table}

\paragraph{Finding 1: The number of HalluCited papers increased sharply in 2025, with EMNLP accounting for over half of these papers.}

Table~\ref{tab:data_stats_HalluCitations} reports the number of papers that contain HalluCitation. The results show that the number of hallucinated papers increased from 20 in 2024 to 275 in 2025. In particular, EMNLP 2025 alone accounts for 154 papers, nearly doubling the number observed at ACL 2025.
Comparing 2024 and 2025, the number of HalluCited papers increased by more than an order of magnitude. The proportion of HalluCited papers also increased from around 0.28\% in 2024 to 2.59\% in 2025, reaching 3.7\% at EMNLP 2025. Examining venue-specific trends in 2025, we find that while workshop papers account for roughly half of the HalluCited papers at NAACL and ACL, nearly 70\% of HalluCited papers at EMNLP 2025 appear in the Main and Findings.
A similar pattern is observed in the proportions of HalluCited papers across tracks in 2025. At NAACL, only workshop papers exceed 3\%, whereas at ACL, the S.I.D. tracks reach 3\%. In contrast, at EMNLP, all tracks exceed 3\%. These results suggest that the impact of HalluCited papers is no longer confined to workshops or side tracks, but is increasingly affecting main tracks.
This finding indicates that the rapid increase in HalluCited papers is less likely to be attributable to deficiencies in specific tracks and instead highlights emerging limitations in the ability of the current review system to detect HalluCitation, revealing its social impact on sustainability and trustworthiness.

\begin{table}[t]
    \centering
    \small
    \setlength{\tabcolsep}{4.5pt}
    \rowcolors{2}{gray!10}{white}
    \begin{tabular}{crrrrrr} \toprule
 & \multicolumn{2}{c}{Candidates} & \multicolumn{2}{c}{HalluCited} & \multicolumn{2}{c}{Hit Rate (\%)} \\
\cmidrule(lr){2-3} \cmidrule(lr){4-5} \cmidrule(l){6-7}
    Freq. &    Num. &  Cum. & Num. &  Cum. &      Num. &     Cum. \\ \midrule
$\geqq$9  &      10 &    10 &   10 &    10 &   100.0\% &  100.0\% \\
       8  &       6 &    16 &    5 &    15 &    83.3\% &   93.8\% \\
       7  &      14 &    30 &   13 &    28 &    92.9\% &   93.3\% \\
       6  &      10 &    40 &    7 &    35 &    70.0\% &   87.5\% \\
       5  &       9 &    49 &    7 &    42 &    77.7\% &   85.7\% \\
       4  &      28 &    77 &   17 &    59 &    60.7\% &   76.6\% \\
       3  &      91 &   168 &   37 &    96 &    40.7\% &   57.1\% \\
       2  &     526 &   694 &   76 &   172 &    14.4\% &   24.8\% \\
       1  &   2,256 & 2,950 &  123 &   295 &    5.45\% &   10.0\% \\
        \bottomrule 
    \end{tabular}
    \caption{Number of candidate papers and detected HalluCited papers. We report the number of papers for each frequency of HalluCitation candidates. Hit Rate denotes the proportion of HalluCited papers within each frequency bin. ``Num.'' indicates the raw count for each bin, and ``Cum.'' represents the cumulative count aggregated from higher frequencies. For papers containing nine or more candidates, all HalluCited papers were successfully detected, and thus, we group them together.}
    \label{tab:data_stats_HalluCitations_number}
\end{table}

\paragraph{Finding 2: Papers containing multiple HalluCitation candidates are highly likely to contain actual HalluCitation.}

Table~\ref{tab:data_stats_HalluCitations_number} shows the detailed counts of candidate papers and the actually detected HalluCited papers. From these results, we find that when a paper contains around four HalluCitation candidates, nearly three out of four papers indeed include HalluCitation. This finding indicates that even simple title matching against reference databases can sufficiently detect HalluCitation.
In contrast, cases with only a few candidates often include noise introduced by OCR or parsing errors, as well as limitations of simple character-level fuzzy matching. 
Although these citations may be erroneously flagged due to parsing failures or high thresholds in fuzzy matching, such mismatches rarely occur consecutively. Therefore, when three or four candidates are detected within a single paper, the paper is considered doubtful and recommended for further verification.

\paragraph{Finding 3: Manual detection is challenging because most HalluCited papers contain only a small number of HalluCitations.}

As shown in Table~\ref{tab:data_stats_HalluCitations_number}, when a paper contains only one or two HalluCitation candidates, the number of detected papers reaches nearly 200, accounting for around two-thirds of all HalluCited papers. This indicates that, rather than papers containing many HalluCitations, the majority of HalluCited papers contain only a small number of HalluCitations.
This also suggests that such HalluCitations are often disguised among otherwise proper citations. At present, only reviewers or readers with strong expertise closely aligned with the cited work may notice these HalluCitations. However, when citations fall outside an area of expertise, manually detecting such hallucinations becomes nearly infeasible within the limited time available, e.g., during the review period.
Moreover, we point out that some HalluCitations may not always be introduced intentionally. For instance, authors may obtain citation information from secondary sources such as Google Scholar\footnote{\url{https://scholar.google.com/}}, reference management tools, or LLM-based recommendation systems, where the entries themselves may already contain hallucinations. We discuss this possibility in detail in Section~\ref{sec:example-based-discussion}.
These HalluCitations highlight that authors often implicitly trust such citations and do not always verify their factuality. Moreover, because such HalluCitations are often embedded among otherwise proper citations, it is preferable to introduce automated detection systems into the validation process.

\section{Further Analysis and Discussions}

\subsection{Trend Analysis}
\label{sec:area-analysis}

\begin{figure}[t]
    \centering
    \includegraphics[width=\linewidth]{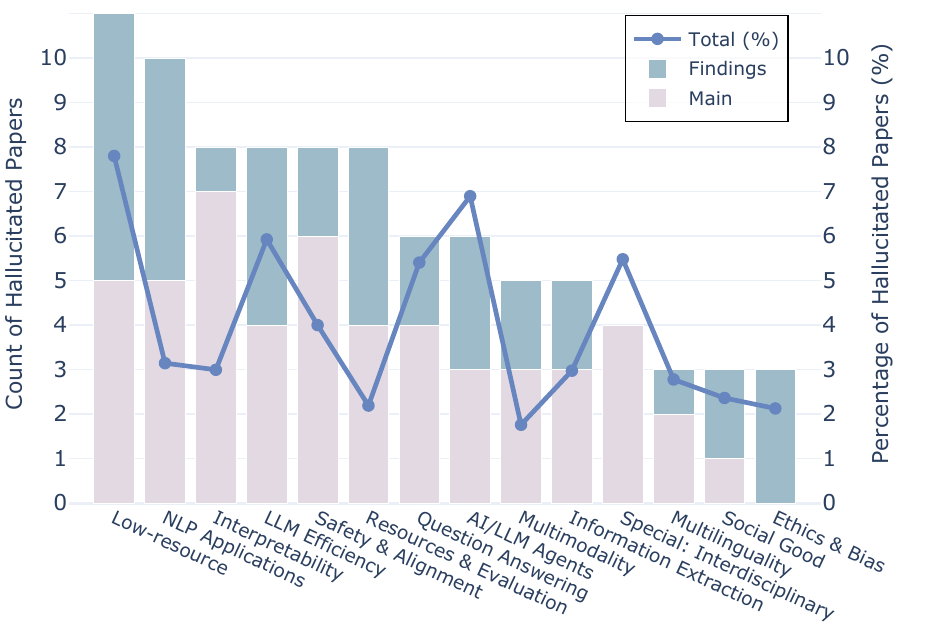}
    \caption{Number and proportion of HalluCited papers by area. We report areas with three or more papers. Area names are abbreviated using the first few words.}
    \label{fig:area_analysis}
\end{figure}

\paragraph{Which areas more frequently include HalluCited papers?}

To analyze trends in the occurrence of HalluCited papers, we examine the research areas in which such papers are accepted.
We focus on EMNLP 2025\footnote{\label{footnote:area-info}[Accessed on Dec. 31st, 2025] Area information is available at \url{https://rrplanning2022-my.sharepoint.com/:x:/g/personal/jrachford_randrplanning\_com/IQCJIl5zmNpUSZUv8j8H0z1TAVaIf1rOpMYhjCllLm703RY} via \url{https://2025.emnlp.org/program}.} and analyze the Main and Findings papers by identifying the assigned research areas for each paper.
Figure~\ref{fig:area_analysis} shows the counts and proportions of HalluCited papers by area. For the official area names and the complete list of areas, please refer to the EMNLP 2025 Call for Papers\footnote{\url{https://2025.emnlp.org/calls/main\_conference\_papers/\#submission-topics}}.
We found that areas such as Low-Resource NLP, LLM Efficiency, and AI/LLM Agents exhibit relatively high counts and proportions of HalluCited papers. In contrast, areas including NLP Applications, Interpretability, and Resources and Evaluation show relatively large absolute counts but lower proportions.
Notably, several areas with higher proportions of HalluCited papers, including Safety and Alignment in LLMs, AI/LLM Agents, and LLM Efficiency, were newly introduced at EMNLP 2025\footnote{\url{https://2025.emnlp.org/track-changes/}}.
This suggests that emerging topics may be difficult to secure qualified reviewers for, which may make rigorous review, such as detecting hallucinations, more challenging.

\paragraph{Which keywords are included in HalluCited papers?}

\begin{figure}[t]
    \centering
    \begin{subfigure}{\linewidth}
        \centering
        \includegraphics[width=0.96\linewidth]{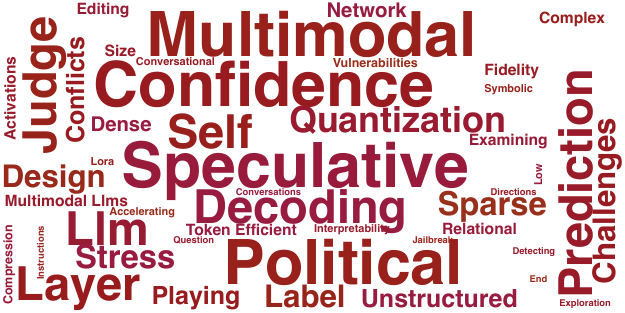}
        \caption{High TF-IDF differences in \textbf{HalluCited papers}.}
        \label{fig:wc-pos}
    \end{subfigure}

    \vspace{0.4em}

    \begin{subfigure}{\linewidth}
        \centering
        \includegraphics[width=0.96\linewidth]{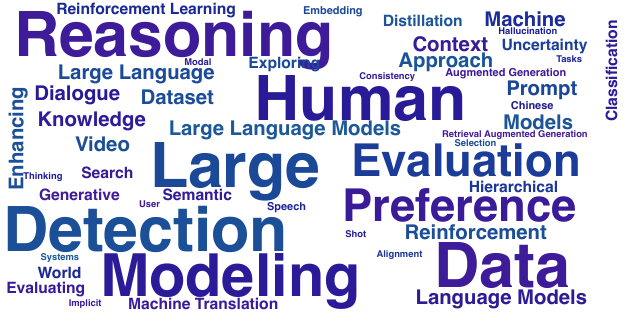}
        \caption{High TF-IDF differences in \textbf{general papers}.}
        \label{fig:wc-neg}
    \end{subfigure}
    \caption{Word clouds based on TF-IDF differences between HalluCited papers and general papers.}
    \label{fig:wordcloud}
\end{figure}

To investigate characteristic terms and topics of HalluCited papers, we compute TF-IDF scores~\cite{manning2008introduction} over paper titles for both general papers and HalluCited papers, focusing on the Main and Findings papers at EMNLP 2025. Next, we compare the two groups by examining terms with large differences in their TF-IDF values. Figure~\ref{fig:wordcloud} shows word clouds~\cite{Mueller_Wordcloud_2023} visualizing terms with high TF-IDF differences for each group.
From these results, we observe that HalluCited papers more frequently contain terms related to topics such as ``Multimodal'', ``Decoding'', and ``Quantization''. In contrast, general papers tend to include terms associated with reinforcement learning-related topics, such as ``Human'', ``Reasoning'', and ``Preference''. In addition, HalluCited papers are more likely to use abbreviations such as ``LLM'' in their titles, whereas general papers more often use the full form, such as ``Large Language Model''.
This suggests that HalluCited papers tend to adopt concise or abbreviated terminology and focus on efficiency-related topics, which is consistent with the results of Figure~\ref{fig:area_analysis}.

\subsection{How about the Peer-Review Process?}
\label{sec:peer-review}

\begin{table}[t]
    \centering
    \small
    \setlength{\tabcolsep}{0.99pt}
    \rowcolors{2}{gray!10}{white}
    \begin{tabular}{crrrrrrrrr} \toprule
 & \multicolumn{2}{c}{2023} & \multicolumn{4}{c}{2024*} & \multicolumn{3}{c}{2025} \\
\cmidrule(lr){2-3} \cmidrule(lr){4-7} \cmidrule(l){8-10}
    Freq. &   Oct. &   Dec. &   Feb. & Apr. &  June & Aug. &  Feb. &   May &  July \\ \midrule
$\geqq$9  &      0 &      0 &      1 &    0 &     0 &    0 &     2 &     6 &     4 \\
       8  &      0 &      0 &      0 &    0 &     0 &    0 &     0 &     1 &     0 \\
       7  &      0 &      0 &      0 &    0 &     0 &    0 &     2 &     3 &     0 \\
       6  &      0 &      0 &      0 &    0 &     0 &    0 &     2 &     3 &     0 \\
       5  &      0 &      0 &      2 &    0 &     3 &    0 &     1 &     5 &     1 \\
       4  &      0 &      0 &      0 &    0 &     0 &    0 &     5 &     6 &     2 \\
       3  &      2 &      0 &      4 &    1 &     3 &    0 &    13 &    14 &     6 \\
       2  &      4 &     18 &     37 &    3 &    39 &    3 &    62 &    60 &    17 \\
       1  &     51 &     87 &    186 &   30 &   226 &   17 &   322 &   237 &    72 \\ \midrule
   cand.  &     57 &    105 &    230 &   34 &   271 &   20 &   409 &   335 &   102 \\  
    total &    270 &    659 &    967 &  227 &   922 &  127 & 1,349 & 1,377 &   371 \\
     \%   &   \textbf{21.1} &   15.9 &   \textbf{23.8} & 15.0 &  \textbf{29.4} & 15.7 &  \textbf{30.3} &  \textbf{24.3} &  \textbf{27.5} \\ \midrule
     \multicolumn{10}{c}{\textsc{Average Review Load}} \\
     meta &    3.9 &    5.0 &    \underline{6.6} &  2.3 &   \underline{5.3} &  2.6 &   \underline{6.5} &   \underline{5.4} &   3.1 \\
   review &    2.2 &    2.4 &    3.7 &  1.9 &   3.1 &  1.8 &   \underline{3.9} &   3.2 &   2.5 \\
\midrule
    \multicolumn{10}{c}{\textsc{Statistics of Available Preprints}} \\
   submit &  1,275 &  2,604 &  4,835 &  881 & 5,813 &  488 & 8,350 & 7,916 & 1,451 \\
     \%   &   21.2 &   25.3 &   20.0 & 25.8 &  15.9 & 26.0 &  16.2 &  17.4 &  25.6 \\ 
        \bottomrule 
    \end{tabular}
    \caption{Number of candidate papers among opted-in preprints in ARR by frequency of HalluCitation candidates. ``cand.'' denotes the number of papers containing at least one HalluCitation candidate, and ``total'' denotes the number of opted-in anonymous preprints. \textbf{Bold} values exceed the average percentage of accepted papers (16.53\%) reported in Table~\ref{tab:data_stats_candidate}. Note that Oct. and Dec. 2024 are excluded because they are not publicly available. Dec. 2023 corresponds to NAACL 2024, Feb. 2024 to ACL 2024, June 2024 to EMNLP 2024, Feb. 2025 to ACL 2025, and May 2025 to EMNLP 2025. We also report the average number of assignments per meta-reviewer and reviewer. \underline{Underlined} values indicate cases exceeding the expected value. We report the total number of submissions and the preprint disclosure rate.}
    \label{tab:arr}
\end{table}

Table~\ref{tab:arr} shows the number of papers containing HalluCitation candidates among opted-in preprints in the ACL Rolling Review (ARR)\footnote{\url{https://aclrollingreview.org/}}, together with reviewer assignment statistics and the disclosure rate among all submissions.
Main and Findings papers are submitted to ARR, and submissions that select the opt-in option are publicly released as preprints. In addition, the ARR Statistics Dashboard\footnote{\url{https://stats.aclrollingreview.org/}} provides aggregate statistics related to ARR. We leveraged these publicly available resources for our analysis.
From Table~\ref{tab:arr}, we observe that for submission iterations linked to conferences since 2024, the proportion of papers containing HalluCitation candidates is high. Nevertheless, when compared with the proportion of accepted papers reported in Table~\ref{tab:data_stats_candidate}, such papers are filtered out to some extent during the peer-review process.
We also observe that papers containing HalluCitation candidates are frequently assigned to meta-reviewers, and in some iterations, such as February 2024, they may be assigned at a rate of at least one paper per reviewer. This highlights the social impact of HalluCited papers as a practical and immediate issue for us.

However, since the preprint disclosure rate is around 20\%, it remains unclear whether such papers are more likely to opt in or opt out. Therefore, we assumed that these papers are uniformly distributed with respect to disclosure. To facilitate retrospective studies and improve transparency in the peer-review process, greater disclosure of review-related information would be important.

Finally, we note that this analysis does not assess the quality of HalluCited papers. It is possible that some low-quality submissions include HalluCitations, and that the other factors already contribute to screening. Nevertheless, given the observed increase in HalluCited papers among accepted papers at EMNLP 2025, filtering such cases is becoming increasingly challenging even with reviewer effort alone. This also suggests the need to implement automated flagging systems to assist reviewer workloads and maintain rigor and quality.

\subsection{Unintentional HalluCitation: Databases Are Often Contaminated}
\label{sec:example-based-discussion}

\begin{figure}[t]
    \centering
    \resizebox{0.92\linewidth}{!}{%
\begin{tcolorbox}[
  width=1.1\linewidth,
  colback=white,
  colframe=black!60,
  boxrule=0.3pt,   %
  left=0.5em, right=0.5em, boxsep=0pt,
  top=0.5em, bottom=0.5em,
  grow to left by=2mm,
  grow to right by=2mm,
]
\begin{description}[leftmargin=*,  itemsep=0.3em, topsep=0.5em, left=0em]
  \item[\underline{Case 1: Cited by \citet{li-etal-2025-frontierscience}}] \ \vspace{0.5em} \\
  \textbf{\textcolor{Red}{\ding{55}} Semantic Scholar~\cite{Ni2024ASE}:} \\
  \citedblock{Xuanfan Ni and Piji Li. 2024a. A systematic evaluation of large language models \uwave{for natural}. ArXiv, abs/2405.10251. {\small\textbf{\href{https://api.semanticscholar.org/CorpusID:261341578}{(Corpus ID: 261341578)}}}} \\
  \textbf{\textcolor{Green}{\ding{51}} arXiv (Formal)~\cite{ni2024systematicevaluationlargelanguage}:} \\
  \citedblock{Xuanfan Ni and Piji Li. 2024b. A systematic evaluation of large language models \uwave{for natural language generation tasks}. Preprint, arXiv:2405.10251.}
  \par\smallskip\hrule\smallskip
  \item[\underline{Case 2: Cited by \citet{lee-etal-2025-quantification}}] \ \vspace{0.5em} \\
  \textbf{\textcolor{Red}{\ding{55}} Google Scholar (non-existent reference):} \\
  \citedblock{Wei Chen, Arjun Kumar, and Lin Yang. 2024. Distraction-based attack prompts: An effective jailbreaking method for llms. Proceedings of the 62nd Annual Meeting of the Association for Computational Linguistics (ACL).  {\small\textbf{\href{https://scholar.google.com/scholar?cluster=10238337808766132122}{(ID: 10238337808766132122)}}}}
  \par\smallskip\hrule\smallskip
  \item[\underline{Case 3: Cited by \citet{ji-etal-2025-pku}}] \ \vspace{0.5em} \\
  \textbf{\textcolor{Red}{\ding{55}} Google Scholar {\small\textbf{\href{https://scholar.google.com/scholar?cluster=3546576102853818527}{(ID: 3546576102853818527)}}}:} \\
  \citedblock{Alex Ray, Joshua Achiam, and Dario Amodei. 2019. Benchmarking safe exploration in deep reinforcement learning. arXiv preprint $\underset{\text{\small\textcolor{Green}{\ding{51}} \citet{fujimoto2019benchmark}}}{\underline{\text{arXiv:1910.01708}}}$, 7(1):2.} \\
  \textbf{\textcolor{Red}{\ding{55}} Semantic Scholar {\small\textbf{\href{https://api.semanticscholar.org/CorpusID:208283920}{(Corpus ID: 208283920)}}}:} \\
  \citedblock{$\underset{\text{\small\textcolor{Red}{\ding{55} Missing First Author, Alex Ray.} }}{\underline{\text{Josh Achiam and Dario Amodei}}}$. 2019. Benchmarking safe exploration in deep reinforcement learning. URL: \url{https://d4mucfpksywv.cloudfront.net/safexp-short.pdf}. \uwave{\textbf{(Expired Link)}}} \\
  \textbf{\textcolor{Green}{\ding{51}} OpenAI (Official)~\cite{Achiam2019BenchmarkingSE}:} \\
  \citedblock{Alex Ray, Josh Achiam, and Dario Amodei. 2019. Benchmarking safe exploration in deep reinforcement learning. URL: \url{https://cdn.openai.com/safexp-short.pdf}} 
\end{description}

\end{tcolorbox}
}
    \caption{Examples of incorrect database entries from Google Scholar and Semantic Scholar. Case 1 shows an incorrect title, Case 2 refers to a non-existent paper, and Case 3 contains inaccurate citation information.}
    \label{fig:incorrect-database-entry}
\end{figure}

HalluCitations may be introduced unintentionally by authors. One common source of such errors is secondary databases, e.g., Google Scholar and Semantic Scholar\footnote{\url{https://www.semanticscholar.org/}}, which may themselves contain incorrect or incomplete entries. Figure~\ref{fig:incorrect-database-entry} shows several cases illustrating this issue.
In Case 1, the paper title is truncated, preventing correct matching with the arXiv database. In Case 2, we confirm that the referenced paper does not exist. There was no corresponding publication by the listed authors at ACL 2024, nor could we find any paper with a similar title, despite the entry being listed in Google Scholar. In Case 3, the citation is associated with an incorrect arXiv ID or is missing an author.

These cases demonstrate that retrieving citation information from secondary sources such as Google Scholar does not guarantee correctness. In particular, as shown in Case 2, entirely hallucinated papers may be registered in such databases, and this citation is in fact referenced by other papers~\cite{zhou2025mllm}. In Case 3, the incorrect arXiv ID has propagated to hundreds of subsequent citations.
These findings suggest that the presence of HalluCitation does not necessarily imply that an AI system generated the citation or that AI tools were used, as erroneous citation entries are frequently observed in these databases. It is plausible that authors manually retrieved citation information from search engines or automatically imported it via reference management tools, such as Paperpile\footnote{\url{https://paperpile.com/}}, thereby inadvertently incorporating incorrect entries.
Furthermore, it is important to address such hallucinations as early as possible. Recent AI-Scientist agents can retrieve citations via APIs from these databases, potentially amplifying the issue and increasing both its scale and complexity over time.

Nevertheless, these cases indicate that authors did not consult the original sources directly and instead relied on secondary sources~\cite{CHURCH_2017}. To mitigate this issue, it is necessary to acknowledge that databases are not error-free and to encourage authors to obtain citation information directly from primary sources such as the ACL Anthology or arXiv. Alternatively, the use of citation normalization tools, such as Ribiber\footnote{\url{https://github.com/yuchenlin/rebiber}}, may help reduce such errors and should be more widely adopted.

\section{Suggestions and Recommendations}

\paragraph{Introducing automatic verification systems for author toolkits and review checks.}

As discussed in Section~\ref{sec:example-based-discussion}, HalluCitations may originate from errors in secondary sources. Therefore, the presence of HalluCitations cannot be directly attributed to the use of LLMs or AI-generated papers alone. However, regardless of their origin, such HalluCitations indicate that authors often fail to carefully verify reference information and do not always consult the original sources.
Therefore, we argue that it is necessary to introduce dedicated HalluCitation detection tools, such as ours or existing LLM-based agents\footnote{\url{https://gptzero.me/news/iclr-2026/}}, and to integrate them into existing toolkits such as ACL pubcheck\footnote{\url{https://github.com/acl-org/aclpubcheck}}. By using these automatic verification systems on both the author and organizer sides, authors can receive early warnings before submission, and organizers can automatically scan submissions at scale. This would enable fairer verification, reduce reviewer workload, and improve overall citation reliability.

\paragraph{Definition of HalluCitation and ensuring traceability in the peer-review process.}

It is important to clearly define what should be considered a HalluCitation, as not all citation errors are equally severe. In this study, we define HalluCitations based on the existence of the cited title and the consistency of key identifiers such as IDs and page numbers. However, minor citation errors have long existed in academic writing~\cite{wang2021science}, and inaccuracies may also originate from secondary sources~\cite{Besan_on_2024}, making their causes complex.
When only a few HalluCitations are identified, it is often sufficient to request correction. Unlike journals, where editors may correct citation information, conference papers place this responsibility largely on authors. Therefore, minor citation errors should not be overly penalized, and providing author toolkits in advance can help prevent many formal mistakes.
A more serious issue arises when it is unclear whether such corrections are reflected in the camera-ready version, highlighting the need for traceability in the revision and peer-review process. Improving it would enable verification of corrections and support rapid analyses of emerging issues such as HalluCitations.

\paragraph{Rethinking the Purpose of Conferences and Peer Review.}
What does ``peer'' review truly mean? We consider reviewers to be hidden co-authors of a paper, whose role is not merely to complete a formal evaluation but to engage in constructive discussion that improves the work. From this perspective, review systems should prioritize sufficient time and manageable workloads that enable meaningful feedback, rather than mechanisms that incentivize rapid review completion under strict time constraints of only a few weeks.
Currently, early release through preprints, such as arXiv, and blog posts has become common. Despite the lack of formal verification through peer review, such outputs are already widely cited~\cite{church-etal-2025-peer}. At the same time, conference reviewing is often less rigorous than journal peer review. If conferences offer neither the immediacy of preprints nor the rigor of journals, their core purpose should be reconsidered. The growing review burden risks undermining conference credibility and raises concerns about long-term sustainability. Without structural change, this situation is likely to worsen.
One possible direction is a community-wide transition toward a mega-journal-style model with asynchronous, rigorous peer review, combined with conference venues that focus on presenting accepted work. Such an evolution could provide a more sustainable review process with sufficient time for verification, while still offering meaningful opportunities for authors to present their research. Originally, the ARR system and the Findings track had the potential to fulfill this role. However, ARR cycles became tightly coupled with conferences, and Findings were eventually positioned as a ``companion''~\cite{findings-2020-findings}. Now, more than five years after the introduction of ARR and Findings, it may be time to reconsider their roles. One possible redefinition is to treat them as journal-equivalent venues, with conference presentations serving as a form of certification or presentation right for accepted work, similar to TMLR. While this would not fundamentally resolve the growth in submission volume, it would at least help preserve credibility. Adapting to such flexible models may become necessary as the community evolves.

\section{Conclusion}
\label{sec:conclusion}

We investigated the presence of HalluCitations in accepted papers at ACL conferences by examining all papers published at ACL, NAACL, and EMNLP in 2024 and 2025. As a result, we identified nearly 300 HalluCited papers, more than half of which appeared at EMNLP 2025.
We argue that the presence of HalluCitations should not be treated as grounds for immediate penalties. As discussed, such errors can arise unintentionally rather than deliberate misconduct by authors. Instead of punitive measures, we emphasize the importance of developing author toolkits and pre-submission checks to proactively prevent such issues.
It is also worth noting that the HalluCited papers were accepted due to their substantive contributions. Accordingly, authors of HalluCited papers should not be penalized post hoc. Since the ACL Anthology allows erratum revisions, it is important to better encourage voluntary corrections\footnote{\url{https://aclanthology.org/info/corrections/}}.
Acceptance should not be treated as a binary reward or punishment, but rather as part of an ongoing process of improving and communicating research.

Moreover, it is important to reaffirm the collaborative nature of peer review. Conferences exist to share work, not to reward perfection under batch review and strict time constraints~\cite{church-2005-last}. \textit{No paper is flawless}, and a healthy research community should remain tolerant of minor errors while actively promoting responsible correction.
The ACL community has long been sustained by self-improvement mechanisms such as Findings papers and ARR. \textit{ACL is not an AI/ML conference.} The NLP community can now take the initiative in shaping peer review and conferences in the LLM era.

\section*{Limitations}

\paragraph{Scope.}
This study focuses on six recent top-tier NLP conferences. Although the scope could be expanded, we limit our analysis for several reasons. First, the number of HalluCited papers is negligible in earlier years, e.g., only two cases at NAACL 2024, and the trend is monotonically increasing, making more retrospective analysis less informative. Second, our primary focus is on the sharp surge observed in 2025, which motivates us to concentrate on the most recent conferences.
Other venues, such as AACL 2025 or the ARR October 2025 cycle, were not publicly available at the time of writing and therefore could not be analyzed, though our findings suggest that a non-trivial number of HalluCitations may also be present there. While extending the analysis to other AI/ML conferences is possible, our goal is to assess the impact within the NLP community specifically.
Overall, despite this limited scope, we believe our analysis is sufficiently comprehensive for the intended purpose, and broader investigations, e.g., other conferences and journals, are left for future work.

\paragraph{Analysis.}
We primarily analyzed accepted papers. First, accepted papers are readily accessible, which makes large-scale analysis feasible. Second, we showed that even accepted papers contain HalluCitations, suggesting that the peer-review process may not always sufficiently verify citation correctness under time constraints and heavy reviewer workloads. From this perspective, analyzing accepted papers is both reasonable and informative.
Nevertheless, we also examined publicly available ARR data and conducted a comprehensive investigation within the limits of available disclosures. However, as discussed in Section~\ref{sec:peer-review}, the transparency of the peer-review process remains limited, and a more thorough analysis is currently infeasible due to restricted access to review materials.
Moreover, rejected papers are generally not publicly available. Although there have been efforts toward data release\footnote{\url{https://arr-data.aclweb.org/}}~\cite{dycke-etal-2023-nlpeer, dycke-etal-2022-yes}, these resources are limited to cases where explicit agreement has been obtained. In addition, the traceability of whether reviewer-requested corrections are reflected in the camera-ready versions remains unclear.
We believe academic papers are the outcome of collaborative work between authors and reviewers. Thus, papers also reflect reviewer feedback. From this perspective, review materials, including rejected papers, should be made publicly available~\cite{church-etal-2025-peer, pmlr-v267-yang25ba, pmlr-v267-kim25am}. Such transparency would facilitate analyses of review quality and outcomes.

\paragraph{Methods.}
We focused on arXiv and the ACL Anthology as solid and well-established data sources. Although extending our analysis to broader domains such as AI, ML, CV, robotics, speech, and journals is an important direction for future work, these domains currently lack a unified and consistently maintained bibliographic infrastructure comparable to that of the ACL community. Since this work focuses on ACL conferences.
Despite this restricted scope, we identify nearly 300 HalluCited papers, indicating that our approach is effective. For implementation, we employ MinerU for OCR, although alternative OCR tools~\cite{wei2025deepseekocr, cui2025paddleocr} and additional engineering improvements are possible. These implementation choices are not our primary contribution. Instead, the key contribution of this work is to quantitatively and comprehensively demonstrate the prevalence and impact of HalluCitations.
Finally, our methodology prioritizes precision, and the reported results should be interpreted as a lower bound. Additional HalluCitations may exist beyond those detected here. We hope that this study will serve as a baseline and motivate follow-up work that improves detection accuracy and coverage.

\section*{Ethical Considerations}

\paragraph{Information.}
In this manuscript, all information reflects the state of publicly available data as of January 4, 2026. Unless explicitly stated otherwise, links, metadata, and dataset contents correspond to what was available at that time. Updates, corrections, or removals after this date may change the information and are not reflected in this study.

\paragraph{Licenses.}
Our primary data sources are the ACL Anthology and OpenReview, both of which are distributed under the Creative Commons Attribution 4.0 International License. We confirm that all additional datasets, dumps, and metadata used in this study are publicly available and permissively licensed. The code and tools developed in this work are authored by us. Human verification was conducted solely by the authors of this paper. Therefore, no licensing or consent issues arise from human annotation in this study.

\paragraph{Discussion on the protection of personally identifiable information and potential harm.}
This study analyzes real published papers. Therefore, the author information of HalluCited papers is inherently observable. However, concealing the identities of HalluCited papers would significantly hinder reproducibility. Moreover, the HalluCitations discussed here can already be identified through public search engines and thus constitute existing public information. Moreover, our analysis is comprehensive and systematic and does not target or single out specific individuals. Accordingly, this study is conducted strictly as academic research.
Importantly, as explicitly stated in Section~\ref{sec:conclusion}, we do not attribute HalluCitations to author misconduct, nor do we argue that authors should be penalized. Rather, we highlight limitations of the current review system, particularly the lack of appropriate author toolkits. Accordingly, this analysis and its claims and purpose are not intended to be offensive or accusatory.
For transparency and reproducibility, we provide a list of HalluCited papers in Section~\ref{sec:list-of-papers}. We do not introduce new annotations or subjective judgments. We only verify and list HalluCitations that already exist in the public record. As an additional safeguard, Table~\ref{tab:list_of_hallucited_papers} in Section~\ref{sec:list-of-papers} does not directly display author names or paper titles of HalluCited papers. Instead, each HalluCited paper is referenced only via links to the corresponding reference entries, and we show one example HalluCitation per paper.
This design balances transparency and reproducibility with the protection of authors. Therefore, our discussion throughout the paper remains neutral and forward-looking, focusing on constructive solutions rather than blame. We have carefully reviewed the ACL Ethics Policy\footnote{\url{https://ethics.aclweb.org/}} and confirm this research complies with its guidelines.

\paragraph{Tool usage.}
We used DeepL, ChatGPT, and Grammarly for translation and grammatical improvement. All original content was written by the authors, and the authors take full responsibility for the final manuscript.
For experiments, we used an NVIDIA A6000 GPU for OCR processing. Other analyses were conducted using a MacBook Pro, spreadsheets, and Google Chrome for literature searches.
Notably, all records related to HalluCitations were managed locally and were not transmitted to any external APIs or services. Accordingly, tool usage and data management were conducted in compliance with ethical considerations.

\bibliography{custom, anthology-1, anthology-2}

\appendix

\section{More Suggestions}

\paragraph{Toward easy verification formats and proper citation practices.}

HalluCitations are often disguised within the reference section, making perceptual detection by human reviewers costly. Although this study relies on OCR-based parsing and manual verification to identify such cases, parsing is not always perfect. Therefore, we believe that reference sections in ACL style formats could be improved to be more machine-friendly to facilitate automatic verification, e.g., by introducing clearer structural delimiters between citations, more explicit formatting of key fields such as titles, and printing URLs rather than embedding them only as hyperlinks.
Moreover, we argue that verifiable identifiers, e.g., clickable links and DOIs, should be mandatory. This would facilitate verification, reduce manual effort, and enable automatic verification.

\paragraph{Eliminating area selection or improving area transparency.}

As discussed in Section~\ref{sec:area-analysis}, the proportion of HalluCited papers varies across research areas, with higher rates in emerging areas. We argue that this discrepancy is partly an artifact of area partitioning. Accordingly, we suggest either moving to a single area or improving transparency by publicly releasing area information for accepted papers.
As cross-area research becomes more popular, fixed area selection can lead to reviewer mismatches and reduced review rigor. Given that modern review systems such as OpenReview support automated reviewer matching, explicit area selection may no longer be necessary. Alternatively, since area information is currently disclosed only partially\footref{footnote:area-info}, archiving it in resources such as the ACL Anthology would help reduce mismatches and support more effective reviews, including the detection of HalluCitations.

\section{List of HalluCited Papers}
\label{sec:list-of-papers}

Table~\ref{tab:list_of_hallucited_papers} shows all identified HalluCited papers and the corresponding evidence, i.e., HalluCitations.
For privacy reasons, we present the information in reference format rather than directly displaying the HalluCited paper details. To identify the corresponding paper, please follow the provided link or refer to the reference section.

\clearpage
\onecolumn

\small{
\renewcommand{\arraystretch}{1.5}
\rowcolors{2}{gray!10}{white}
\setlength{\tabcolsep}{4pt}

\begin{xltabular}{\linewidth}{ll
>{\raggedright\arraybackslash}p{0.2\linewidth}
>{\raggedright\arraybackslash}p{0.2\linewidth}
>{\raggedright\arraybackslash}X
}

\hiderowcolors
\caption{List of hallucinated papers, i.e., HalluCited Papers, and their corresponding HalluCitations.}
\
\label{tab:list_of_hallucited_papers} \\
\showrowcolors

\hiderowcolors
\toprule
 & \textsc{Event ID} & \textsc{Paper ID} & \textsc{HalluCited Paper} & \textsc{Example of HalluCitation} \\
\midrule
\showrowcolors
\endfirsthead

\hiderowcolors
\toprule
 & \textsc{Event ID} & \textsc{Paper ID} & \textsc{HalluCited Paper} & \textsc{Example of HalluCitation} \\
\midrule
\showrowcolors
\endhead

\hiderowcolors
\midrule
\multicolumn{5}{r}{\textsc{Continued on next page}} \\
\midrule
\showrowcolors
\endfoot

\hiderowcolors
\midrule
\multicolumn{5}{r}{\textsc{Concluded}} \\
\bottomrule
\showrowcolors
\endlastfoot

1 & naacl-2024 & 2024.naacl-short.63 & \citet{scaria-etal-2024-instructabsa} & Himanshu Gupta, Kevin Scaria, Swaroop Mishra, and Chitta Baral. 2024b. Beyond the data bottleneck: Optimizing instruction tuning with difficulty-based exemplar selection. ArXiv preprint. \\
2 & naacl-2024 & 2024.trustnlp-1.1 & \citet{adilazuarda-2024-beyond} & Daphne Ippolito, Daniel Duckworth, Chris CallisonBurch, and Douglas Eck. 2020. Discriminating between human-produced and machine-generated text: A survey. arXiv preprint arXiv:2012.03358. \\
3 & acl-2024 & 2024.acl-long.560 & \citet{siska-etal-2024-examining} & Keisuke Sakaguchi, Ronan Le Bras, Chandra Bhagavatula, and Yejin Choi. 2023. An adversarial winograd schema challenge at scale. arXiv preprint arXiv:2305.06300. \\
4 & acl-2024 & 2024.acl-long.611 & \citet{zheng-etal-2024-self} & Chun-Hung Yeh, Anuradha Welivita, and Pearl Pu Faltings. 2015. A dialogue dataset containing emotional support for people in distress. arXiv preprint arXiv:1503.08895. \\
5 & acl-2024 & 2024.acl-long.768 & \citet{chen-etal-2024-logogramnlp} & Mengjie Fang, Linlin Xu, and Pascale Fung. 2020. Unsupervised cross-lingual transfer learning for contextualized word embeddings. In Proceedings of the 2020 Conference on Empirical Methods in Natural Language Processing (EMNLP), pages 2616–2629. \\
6 & acl-2024 & 2024.arabicnlp-1.24 & \citet{khondaker-etal-2024-benchmarking} & Isabel Moreno and Wei Zhang. 2024. Evaluating multilingual models on nlp tasks in arabic. Computational Linguistics, 50(3):425–445. \\
7 & acl-2024 & 2024.arabicnlp-1.45 & \citet{shah-etal-2024-mememind} & Naoya Inoue, Pontus Stenetorp, and Kentaro Inui. 2021. Interplay between preferences and machine learning in language model fine-tuning. arXiv preprint arXiv:2110.08413. \\
8 & acl-2024 & 2024.findings-acl.241 & \citet{holt-etal-2024-perceptions} & Junjie Li, Jieyu Wu, and Richard Socher. 2021. Understanding code-switching in language models. arXiv preprint arXiv:2109.04278. \\
9 & acl-2024 & 2024.wassa-1.18 & \citet{alhamed-etal-2024-monitoring} & David Benetka, Alicia Moreno-Moral, Lorena RomeroFombuena, and Juan Lopez-Gazpio. 2020. An annotation scheme for mental health discussions in social media. In International Conference on Computational Linguistics (Proceedings of the Conference: Long Papers, 2020), pages 2617–2627. Association for Computational Linguistics. \\
10 & emnlp-2024 & 2024.conll-1.8 & \citet{mundra-etal-2024-empirical} & Adam Alabi, Saleha Nawaz, and Vincent Ng. 2022. Alabi: A light-weight approach for multilingual biomedical language models. In Proceedings of the 2022 Conference on Empirical Methods in Natural Language Processing, pages 9717–9727. \\
11 & emnlp-2024 & 2024.emnlp-demo.11 & \citet{sheth-etal-2024-commentator} & Stephan Druskat, Ulrike Gut, Nils Reiter, Stefan Schweter, and Manfred Stede. 2014. Atomic: An open-source tool for working with anaphora in multiple languages. In Proceedings of the 2014 Conference on Empirical Methods in Natural Language Processing: System Demonstrations, pages 71–76. \\
12 & emnlp-2024 & 2024.emnlp-industry.112 & \citet{malik-etal-2024-pearl} & Jiajia Liu, Mengyuan Yang, Yankai Yu, Haixia Xu, Kang Li, and Xiaobo Zhou. 2023a. Enhancing customer experience with ai-driven conversational agents. arXiv preprint arXiv:2306.04325. \\
13 & emnlp-2024 & 2024.emnlp-main.345 & \citet{xiao-etal-2024-toxicloakcn} & Nelson F Liu, Tony Wu, Duane S Boning, and Tanmoy Choudhury. 2020b. AI bug detector: Adversarial input detection for natural language processing models. In Proceedings of the 2020 Conference on Empirical Methods in Natural Language Processing: System Demonstrations, page 187–196. \\
14 & emnlp-2024 & 2024.emnlp-main.464 & \citet{wang-etal-2024-fedkim} & Yu Gu, Robert Tinn, Hao Cheng, Youzheng Ben, Zhuozhao Liu, Jingqi Zhou, Michael Wang, Shizhuo Wang, Hongfang Zhou, and Yanshan Shen. 2021. Biomedgpt: A large-scale biomedical generative pretrained transformer for biomedical text mining. arXiv preprint arXiv:2104.07497. \\
15 & emnlp-2024 & 2024.emnlp-main.617 & \citet{wahle-etal-2024-paraphrase} & Dominik Meier, Jan Philip Wahle, Terry Ruas, and Bela Gipp. 2024. A human study on atomic paraphrase type generation. arXiv. \\
16 & emnlp-2024 & 2024.findings-emnlp.232 & \citet{zhang-etal-2024-mabc} & Yi Chen et al. 2023. Empowering cloud rca with augmented large language models. arXiv preprint arXiv:2311.00000. \\
17 & emnlp-2024 & 2024.findings-emnlp.813 & \citet{li-etal-2024-drattack} & Ruochen Wang, Ting Liu, Cho-jui Hsieh, and Boqing Gong. 2023. DPO-DIFF:on Discrete Prompt Optimization for text-to-image DIFFusion modelsgenerating Natural Language Adversarial Examples. Preprint, arXiv:2311.07998. \\
18 & emnlp-2024 & 2024.findings-emnlp.912 & \citet{sonkar-etal-2024-student} & Yoav Goldberg. 2022. Assessing claims about large language models. arXiv preprint arXiv:2212.09273. \\
19 & emnlp-2024 & 2024.findings-emnlp.981 & \citet{zafar-etal-2024-medlogic} & Xinyu Wang, Tao Sun, Deqing Zou, Wei Wu, and Jiawei Han. 2021. Logic-guided data augmentation and regularization for consistency learning. arXiv preprint arXiv:2104.04379. \\
20 & emnlp-2024 & 2024.mrl-1.20 & \citet{skianis-etal-2024-leveraging} & Shaoxiong Ji, Yanzhao Zhang, Leilei Sun, and Jia Wang. 2022. Mentalbert: A pretrained language model for mental healthcare. In Proceedings of the 2022 Conference on Empirical Methods in Natural Language Processing (EMNLP). \\
21 & naacl-2025 & 2025.americasnlp-1.4 & \citet{alvarez-c-etal-2025-advancing} & Manuel Mager, Arturo Oncevay, Annette Rios, Jamshidbek Mirzakhalov, and Katharina Kann. 2023. The role of computational linguistics in indigenous language revitalization: Challenges and opportunities. In Proceedings of the 1st Workshop on Computation for Indigenous Languages (C3NLP), pages 23–31. \\
22 & naacl-2025 & 2025.clpsych-1.23 & \citet{antony-schoene-2025-retrieval} & Ilias Chalkidis, Ion Androutsopoulos, and Nikolaos Aletras. 2022. An empirical study on neural methods for legal judgment prediction. In Proceedings of the 2022 Conference on Empirical Methods in Natural Language Processing, pages 4783–4798. Association for Computational Linguistics. \\
23 & naacl-2025 & 2025.dravidianlangtech-1.117 & \citet{noor-etal-2025-celestia} & G. Shimi et al. 2024. Language identification for dravidian languages: A crucial step for fake news detection in multilingual settings. TBD. \\
24 & naacl-2025 & 2025.dravidianlangtech-1.118 & \citet{selvamurugan-2025-dravlingua} & Anil Kumar, Ajay Kumar Ojha, Shervin Malmasi, and Marcos Zampieri. 2021. Benchmarking aggression identification in social media for low-resource languages. In Proceedings of the Workshop on Online Abuse and Harms (ACL). \\
25 & naacl-2025 & 2025.dravidianlangtech-1.38 & \citet{s-etal-2025-ksk} & Simran Khanuja, Kaustav Dey, El Moatez Billah Karim Nagoudi, et al. 2020. A new dataset and strong baselines for the detection of code-mixed offensive language. In Proceedings of the 2020 Conference on Empirical Methods in Natural Language Processing (EMNLP), pages 1719–1726. \\
26 & naacl-2025 & 2025.dravidianlangtech-1.43 & \citet{chauhan-kumar-2025-mnlp-dravidianlangtech} & Douwe Kiela, Mohamed Elhoseiny, Lin Zhang, Marco Baroni, and Geoffrey Hinton. 2019. Supervised multimodal hashing for scalable cross-modal retrieval. In Proceedings of the 2019 Conference on Empirical Methods in Natural Language Processing (EMNLP), pages 1740–1751. \\
27 & naacl-2025 & 2025.dravidianlangtech-1.61 & \citet{k-etal-2025-nlp} & F. Wu and M. Dredze. 2019. Social media as a sensor of public opinion. Proceedings of the 2019 Conference on Empirical Methods in Natural Language Processing (EMNLP 2019), 2019:1001–1010. \\
28 & naacl-2025 & 2025.dravidianlangtech-1.62 & \citet{vaidyanathan-etal-2025-nlp} & Jiho Park, Jihyung Shin, Sangyoun Lee, and Changhyun Seo. 2018. A survey of hate speech detection: Data, methods, and challenges. In Proceedings of the 27th International Conference on Computational Linguistics (COLING 2018), pages 385–395. International Committee on Computational Linguistics. \\
29 & naacl-2025 & 2025.dravidianlangtech-1.7 & \citet{sathvik-etal-2025-team} & M. H. Al-Adhaileh and F. W. Alsaade. 2022. Bidirectional long-short term memory (bilstm) networks for fake review detection: A comparative study. Springer. \\
30 & naacl-2025 & 2025.dravidianlangtech-1.88 & \citet{achamaleh-etal-2025-cic} & Anna Kolesnikova and Sergey Ivanov. 2023. Exploring multilingual text representations with transformer models. Transactions of the ACL, 11:212–230. \\
31 & naacl-2025 & 2025.dravidianlangtech-1.91 & \citet{aftahee-etal-2025-cuet} & R. Raja, A. Kumar, and S. Joseph. 2023. Fake news detection in low-resource languages: Challenges and advancements. Computational Linguistics Review, 15(2):123–137. \\
32 & naacl-2025 & 2025.dravidianlangtech-1.97 & \citet{sathvik-etal-2025-detection} & Simran Khanuja, Sandipan Dandapat, Ritesh Kumar, Sunayana Sitaram, K. P. Soman, and Anup Kumar. 2021. Mahanlp: Towards indic language understanding using bert models for hindi, marathi, and kannada. arXiv preprint arXiv:2106.07469. \\
33 & naacl-2025 & 2025.findings-naacl.132 & \citet{singh-etal-2025-fidelity} & W. Liu, X. Li, W. Yang, Y. Lin, X. Liu, S. Wang, C. Xie, L. Xu, W. Zhuang, X. Zhao, and L. Li. 2020. Minilm: Deep self-attention distillation for tiny transformers. arXiv preprint arXiv:2002.10957. \\
34 & naacl-2025 & 2025.findings-naacl.223 & \citet{zhu-etal-2025-grait} & Krishnateja Killamsetty and Rishabh Iyer. 2021. Subset selection with gradient matching. arXiv preprint arXiv:2103.00123. \\
35 & naacl-2025 & 2025.findings-naacl.230 & \citet{raihan-etal-2025-mojobench} & OpenAI. 2024. Gpt-4 omni: A comprehensive multimodal model for language, vision, and beyond. arXiv preprint arXiv:2408.01234. \\
36 & naacl-2025 & 2025.findings-naacl.257 & \citet{magomere-etal-2025-finnli} & Chaitanya Shivade et al. 2019. Mednli-a natural language inference dataset for the clinical domain. In Proceedings of the 2018 Conference on Empirical Methods in Natural Language Processing, Brussels, Belgium. Association for Computational Linguistics, pages 1586–1596. \\
37 & naacl-2025 & 2025.findings-naacl.275 & \citet{song-etal-2025-echoes} & M Honnibal and I Montani. 2017. spacy 2: Natural language understanding with bloom embeddings, convolutional neural networks and incremental parsing. neural machine translation. In Proceedings of the Association for Computational Linguistics (ACL), pages 688–697. \\
38 & naacl-2025 & 2025.findings-naacl.279 & \citet{basher-etal-2025-bntts} & Y. Xu et al. 2023. Cross-lingual transfer for lowresource text-to-speech. In Proceedings of the 2023 Conference on Empirical Methods in Natural Language Processing, pages 510–520. Association for Computational Linguistics. \\
39 & naacl-2025 & 2025.findings-naacl.349 & \citet{gupta-etal-2025-selective} & Zijun Zhang, Yue Wu, Hao Guan, Xinlei Chen, and Yue Zhang. 2023. Continual learning with transformers: Challenges and solutions. arXiv preprint arXiv:2302.13713. \\
40 & naacl-2025 & 2025.findings-naacl.418 & \citet{nguyen-etal-2025-large} & Danny Hernandez, Jared Kaplan, Tom Henighan, Tom Brown, Jack Clark, Preetum Nakkiran, Catherine Olsson, and Afshin Rahimi. 2023. Measuring data contamination in large language models. arXiv preprint arXiv:2303.13375. \\
41 & naacl-2025 & 2025.findings-naacl.452 & \citet{wu-etal-2025-grapheval36k} & Abhijit Srivastava, Nathan Major, Jasmine D. Hernandez, and et al. 2022. Beyond the imitation game: Quantifying and extrapolating llm capabilities with big-bench. arXiv preprint arXiv:2206.04615. \\
42 & naacl-2025 & 2025.findings-naacl.58 & \citet{quancai-etal-2025-discomp} & Xiang Jiang et al. 2023c. Mistral-7b-instruct-v0.2: An efficient large language model for instruction following. arXiv preprint arXiv:2309.00001. \\
43 & naacl-2025 & 2025.knowledgenlp-1.18 & \citet{spangher-etal-2025-novel} & Tushar Khot, Ashish Sabharwal, and et al. 2023. Decomposition-driven reasoning in language models. arXiv preprint arXiv:2304.xxxxx. \\
44 & naacl-2025 & 2025.knowledgenlp-1.20 & \citet{pattnayak-etal-2025-hybrid} & et al. Zhao, W. 2020. A retrieval-augmented encoder-decoder for knowledge-intensive nlp tasks. In Proceedings of the 58th Annual Meeting of the Association for Computational Linguistics (ACL). \\
45 & naacl-2025 & 2025.loresmt-1.12 & \citet{raja-vats-2025-parallel} & Kalika Bali, Jatin Sharma, Monojit Choudhury, and Yogarshi Vyas. 2014. Code-mixing: A challenge for language identification in the indian context. In Proceedings of the First Workshop on Computational Approaches to Code Switching (EMNLP), pages 13–23. \\
46 & naacl-2025 & 2025.naacl-demo.39 & \citet{hiray-kovatchev-2025-storybranch} & T. Lee et al. 2023. Bivdiff: A training-free framework for general-purpose video synthesis. arXiv preprint arXiv:2302.02918. \\
47 & naacl-2025 & 2025.naacl-industry.68 & \citet{jararweh-etal-2025-protein2text} & Haotian Liu, Chunyuan Lin, Fangyun Zeng, and et al. 2023a. Llava: Large language and vision assistant. arXiv preprint arXiv:2304.08485. \\
48 & naacl-2025 & 2025.naacl-long.182 & \citet{chen-etal-2025-benchmarking} & Ebtesam Almazrouei, Hamza Alobeidli, Abdulaziz Alshamsi, Alessandro Cappelli, Ruxandra Cojocaru, Merouane Debbah, Etienne Goffinet, Daniel Heslow, Julien Launay, Quentin Malartic, et al. 2023. Falcon-40b: an open large language model with stateof-the-art performance. Findings of the Association for Computational Linguistics: ACL, 2023:10755–10773. \\
49 & naacl-2025 & 2025.naacl-long.363 & \citet{qian-etal-2025-beyond} & Zhengbao Jiang, Frank F Xu, Jun Araki, and Graham Neubig. 2020. Can prompt-based learning be practical for low-resource tasks? arXiv preprint arXiv:2001.07676. \\
50 & naacl-2025 & 2025.naacl-long.371 & \citet{wysocka-etal-2025-syllobio} & Jakob Prange, Khalil Mrini, and Noah A. Smith. 2023. Challenges and opportunities in nlp for systematic generalization: Beyond compositionality. In Proceedings of the 61st Annual Meeting of the Association for Computational Linguistics. Association for Computational Linguistics. \\
51 & naacl-2025 & 2025.naacl-long.620 & \citet{chen-etal-2025-mastering} & Jiawei Liu, Yifeng Ding, Naman Jain, Harm de Vries, Leandro von Werra, Arjun Guha, Lingming Zhang, and Yuxiang Wei. 2024a. Starcoder2-instruct: Fully transparent and permissive self-alignment for code generation. Preprint, arXiv:2307.08701. \\
52 & naacl-2025 & 2025.naacl-long.72 & \citet{hu-etal-2025-large-language} & Mor Geva, Yoav Goldberg, and Jonathan Berant. 2021. Strategyqa: A question answering benchmark for reasoning about strategies. In Proceedings of the 2021 Conference on Empirical Methods in Natural Language Processing. \\
53 & naacl-2025 & 2025.naacl-short.79 & \citet{kumar-etal-2025-mixrevdetect} & Sankha Bhattacharjee et al. 2023. Adversarial learning for robust ai-generated text detection. arXiv preprint arXiv:2304.07812. \\
54 & naacl-2025 & 2025.nlp4dh-1.52 & \citet{greschner-klinger-2025-fearful} & Chiara Leoni, Mauro Coccoli, Ilaria Torre, and Gianni Vercelli. 2018. Your paper title here. In Proceedings of the Fifth Italian Conference on Computational Linguistics (CLiC-it 2018), Torino, Italy. Accademia University Press. \\
55 & naacl-2025 & 2025.wnu-1.10 & \citet{li-etal-2025-beyond-llms} & Xingxing Lu, Yuning Mao, and Jason Wei. 2023. Autoeval: Llm-based automatic evaluation framework for text summarization. In Findings of ACL. \\
56 & acl-2025 & 2025.acl-demo.16 & \citet{she-etal-2025-token} & Wenhao Zheng and 1 others. 2025a. Citer: Confidencebased token routing for collaborative inference with large language models. arXiv preprint arXiv:2503.01013. \\
57 & acl-2025 & 2025.acl-demo.25 & \citet{zheng-etal-2025-spatialwebagent} & Wei Zhang, Pengfei Liu, and Qiang Shen. 2020. Handling ambiguities in geographic texts using nlp techniques. Transactions in GIS, 24(3):519–538. \\
58 & acl-2025 & 2025.acl-demo.44 & \citet{yang-etal-2025-docagent} & Ziniu Wu, Cheng Liu, Jindong Zhang, Xinyun Li, Yewen Wang, Jimmy Xin, Lianmin Zhang, Eric Xing, Yuxin Lu, and Percy Liang. 2023. Autogen: Enabling next-generation multi-agent communication with language models. arXiv preprint arXiv:2309.07864. \\
59 & acl-2025 & 2025.acl-industry.30 & \citet{fu-etal-2025-model} & Yihan Chen, Dongkuan Zhang, Xiang Wang, Yifan Yang, and Heng Wang. 2023. Dare: ldirect parameter editing for adaptive mode reconfiguration. arXiv preprint arXiv:2310.09570. \\
60 & acl-2025 & 2025.acl-industry.54 & \citet{zhu-etal-2025-plangpt} & Yijie Deng, He Zhu, Wen Wang, Minxin Chen, Junyou Su, and Wenjia Zhang. 2025. Urban planning bench: A comprehensive benchmark for evaluating urban planning capabilities in large language models. †Equal contribution. *Corresponding author: wenjiazhang@tongji.edu.cn. \\
61 & acl-2025 & 2025.acl-long.1034 & \citet{zhou-di-eugenio-2025-veracity} & Jane Doe and Wei Chen. 2023. Debiasing reasoning in language models: A quantitative study. Proceedings of the 2023 Conference on Empirical Methods in Natural Language Processing. \\
62 & acl-2025 & 2025.acl-long.1369 & \citet{chen-etal-2025-inner} & Mia Xu Chen, Orhan Firat, Ankur Bapna, Melvin Johnson, and Wolfgang Macherey. 2024a. Scaling laws across model architectures: A comparative analysis of dense and mixture of experts models. arXiv preprint arXiv:2410.05661. \\
63 & acl-2025 & 2025.acl-long.1409 & \citet{frydenlund-2025-language} & Jason Lee, Elman Mansimov, and Kyunghyun Cho. 2018. Deterministic non-autoregresinterpreting gpt: the logit lenssive neural sequence modeling by iterative refinement. In Proceedings of the 2018 Conference on Empirical Methods in Natural Language Processing, pages 1173–1182, Brussels, Belgium. Association for Computational Linguistics. \\
64 & acl-2025 & 2025.acl-long.1435 & \citet{jia-etal-2025-task} & Mei Li et al. 2024. Securing tool use in llm agents: Challenges and strategies. arXiv preprint arXiv:2402.03014. \\
65 & acl-2025 & 2025.acl-long.1439 & \citet{siingh-etal-2025-getreason} & Wei Xu, Shujie Liu, Minlie Huang, Si Wei, and Hang Li. 2018. A survey on hallucination in neural machine translation. arXiv preprint arXiv:1803.02577. Accessed: 2025-02-14. \\
66 & acl-2025 & 2025.acl-long.1554 & \citet{klisura-etal-2025-multi} & Dorottya Demeszky, Weijie Gong, and Diyi Yang. 2019. Analyzing bias and framing in news articles on artificial intelligence: The case of sentiment analysis. In Proceedings of the 2019 Annual Conference on Empirical Methods in Natural Language Processing, pages 1655–1661. \\
67 & acl-2025 & 2025.acl-long.1580 & \citet{spangher-etal-2025-newsinterview} & Minjun Chang et al. 2023. Self-reflection with generative agents. arXiv preprint arXiv:2311.09214. \\
68 & acl-2025 & 2025.acl-long.248 & \citet{lee-etal-2025-quantification} & Wei Chen, Arjun Kumar, and Lin Yang. 2024. Distraction-based attack prompts: An effective jailbreaking method for llms. Proceedings of the 62nd Annual Meeting of the Association for Computational Linguistics (ACL). \\
69 & acl-2025 & 2025.acl-long.272 & \citet{meng-etal-2025-stigma} & Chen et al. 2023a. Nl2fol: Natural language to firstorder logic. arXiv preprint arXiv:2304.09102. \\
70 & acl-2025 & 2025.acl-long.340 & \citet{li-etal-2025-drift} & openai. 2024. Learning to reason with llm. Preprint, arXiv:2406.19949. \\
71 & acl-2025 & 2025.acl-long.421 & \citet{zhu-etal-2025-multiagentbench} & Xu Sun, Xiaoya Zhang, Yicheng Feng, Shiyang Wang, Shuming Ma, Jiuding He, Zhixu Zhang, Yuxian Gu, Yi Xu, Hao Zhou, and Zhiyuan Liu. 2023. A systematic capability evaluation of large vision-language models. arXiv preprint arXiv:2305.16372. \\
72 & acl-2025 & 2025.acl-long.491 & \citet{alsagheer-etal-2025-lawyer} & Hao Zhong, Jieyu Tang, Tianyang Xu, et al. 2020. Does nlp benefit legal system? a case study of document representation. arXiv preprint arXiv:2005.01647. \\
73 & acl-2025 & 2025.acl-long.922 & \citet{mohammad-2025-words} & Jan Philip Wahle, Krishnapriya Vishnubhotla, Bela Gipp, and Saif M. Mohammad. 2025. Affect, body, cognition, demographics, and emotion: The abcde of text features for computational affective science. arXiv. \\
74 & acl-2025 & 2025.acl-long.979 & \citet{chang-etal-2025-automixer} & Xinyu Li, Rahul Gupta, and Emily Brown. 2024. Datacomplm: Benchmarking data quality in large language models. arXiv preprint arXiv:2402.23456. \\
75 & acl-2025 & 2025.acl-srw.40 & \citet{wu-etal-2025-libvulnwatch} & [first names omitted for brevity] Zhang and colleagues. 2024. Metagpt: Meta programming sota autonomous multi-agent cooperative llm workflows. arXiv preprint arXiv:2404.14496. \\
76 & acl-2025 & 2025.acl-srw.56 & \citet{pulido-etal-2025-speak} & Sabrina J. Mielke, Tal Linzen, and Jason Eisner. 2021. What kind of knowledge is captured by contextualized word representations? In Proceedings of the 59th Annual Meeting of the Association for Computational Linguistics, pages 1250–1265. \\
77 & acl-2025 & 2025.acl-srw.73 & \citet{sakunkoo-sakunkoo-2025-lost} & Christo Kirov, John Sylak-Glassman, Ryan Que, David Yarowsky, Jason Eisner, and Ryan Cotterell. 2018. Universal morphological inflection generation using a multilingual dataset. Proceedings of the 2018 Conference of the North American Chapter of the Association for Computational Linguistics (NAACL), pages 52–62. \\
78 & acl-2025 & 2025.acl-srw.85 & \citet{dzhubaeva-etal-2025-unstructured} & Y. Wang, H. Li, and X. Zhang. 2024. Consistency of personality traits in quantized role-playing dialogue agents. In Proceedings of the 2024 Conference on Empirical Methods in Natural Language Processing: Industry Track, pages 123–130. \\
79 & acl-2025 & 2025.acl-srw.98 & \citet{kumar-nolbaria-2025-bridging} & Raghuraj Sailesh Shah, Ankur Anand, Srini Chodisetti, Ankit Gupta, Raj Sanjay Patel, Sameer Patel, and Manish Gupta. 2022. FinservNLP: A library of financial shared tasks and benchmarks. In Proceedings of the Third Workshop on Economics and Natural Language Processing, pages 144–150. Association for Computational Linguistics. \\
80 & acl-2025 & 2025.africanlp-1.1 & \citet{akpobi-2025-yankari} & Adewale Oladipo et al. 2023. Wura: A multilingual dataset of african languages. In Proceedings of the 61st Annual Meeting of the Association for Computational Linguistics. \\
81 & acl-2025 & 2025.africanlp-1.15 & \citet{ismail-etal-2025-retrieval} & J. Sohn and 1 others. 2024. Rationale-guided rag for medical question answering. arXiv preprint arXiv:2411.00300. \\
82 & acl-2025 & 2025.africanlp-1.30 & \citet{pandya-etal-2025-swahili} & M. Martin et al. 2022. Swahbert: Enhancing swahili nlp with pretrained transformers. In Proc. EACL. \\
83 & acl-2025 & 2025.argmining-1.30 & \citet{el-baff-etal-2025-criticalbrew} & Yiqiu Xu, Alessio Frosini, Mattia Vanni, Anisa Rula, and Roberto Navigli. 2024. Frase: Frame-based semantic enrichment for sparql query generation. arXiv preprint arXiv:2503.22144. \\
84 & acl-2025 & 2025.bea-1.66 & \citet{rezayi-etal-2025-automated} & Shuai Bai, Keqin Chen, Xuejing Liu, Jialin Wang, Wenbin Ge, Sibo Song, Kai Dang, Peng Wang, Shijie Wang, Jun Tang, Humen Zhong, Yuanzhi Zhu, Mingkun Yang, Zhaohai Li, Jianqiang Wan, Pengfei Wang, Wei Ding, Zheren Fu, Yiheng Xu, Jiabo Ye, Xi Zhang, Tianbao Xie, Zesen Cheng, Hang Zhang, Zhibo Yang, Haiyang Xu, and Junyang Lin. 2024. Qwen2.5: The next generation of qwen large language models. arXiv preprint arXiv:2407.10671. \\
85 & acl-2025 & 2025.bea-1.73 & \citet{nebhi-etal-2025-end} & Elizabeth Mayfield and Alan W. Black. 2020. Automated scoring of written essays with transformer models. In Proc. BEA at ACL 2020. \\
86 & acl-2025 & 2025.bea-1.86 & \citet{jain-rengarajan-2025-emergent} & Tianyu Zhang, Sayak Basu, Douwe Kiela, and Oriol Vinyals. 2024. Cascade-aware training and inference for more resource-efficient language models. arXiv preprint arXiv:2405.12345. \\
87 & acl-2025 & 2025.bionlp-1.6 & \citet{kumar-basu-2025-adabiobert} & Ilias Chalkidis, Michail Fergadiotis, Roger Stradling, Nikolaos Pappas, and Prodromos Malakasiotis. 2020. Transformer-based models for legal and biomedical document classification: A comparative study. In Proceedings of ACL. \\
88 & acl-2025 & 2025.bionlp-share.22 & \citet{evgin-etal-2025-metninozu} & Qingxiu Jia, Qipeng Xu, Weiting Yu, Yitong Duan, JianYun Nie, and Zhiyuan Liu. 2022. Alignscore: Evaluating factual consistency with contextual alignment. In Proceedings of the 2022 Conference on Empirical Methods in Natural Language Processing (EMNLP). \\
89 & acl-2025 & 2025.bionlp-share.33 & \citet{xu-etal-2025-team} & Yang Liu, Ani Nenkova, and Kathleen McKeown. 2022. Incorporating controllability in neural abstractive summarization. In Proceedings of the 60th Annual Meeting of the ACL, pages 876–890. ACL. \\
90 & acl-2025 & 2025.bionlp-share.34 & \citet{phasook-etal-2025-vereafine} & Sahil Dhuliawala, Raghav Gupta, and Shashi Narayan. 2023. Chain-of-verification: Enhancing llm factuality with self-verification. In NAACL. \\
91 & acl-2025 & 2025.findings-acl.114 & \citet{gupta-etal-2025-carmo} & Yao Du, Kaitao Qian, Sanmi Koyejo, Zheng Zheng, and Chao Zhang. 2023. Alpaca: A strong, replicable instruction-following model. arXiv preprint arXiv:2303.16199. \\
92 & acl-2025 & 2025.findings-acl.1140 & \citet{cui-etal-2025-heuristic} & Xiaoyu Li, Wei Zhang, Jiahui Chen, and Hongwei Liu. 2024c. Iris: Benchmarking large language models for information retrieval tasks. arXiv preprint arXiv:2401.12345. \\
93 & acl-2025 & 2025.findings-acl.1143 & \citet{das-etal-2025-dpo} & Raphael Rafailov, Orion Redwood, et al. 2023. Direct preference optimization: You don’t need rewards to finish rlhf. arXiv preprint arXiv:2305.11517. Preprint, arXiv:2305.11517. \\
94 & acl-2025 & 2025.findings-acl.1156 & \citet{zafrir-etal-2025-fastdraft} & Shashi Narayan, Shay B. Cohen, and Mirella Lapata. 2018. Xsum: A new dataset for abstractive summarization of news articles. In Proceedings of the 2018 Conference on Empirical Methods in Natural Language Processing (EMNLP), pages 701–711. \\
95 & acl-2025 & 2025.findings-acl.1184 & \citet{garg-etal-2025-just} & William Berrios, Yonglong Tian, Chiraag Lala, Alec Parrish, and Trevor Darrell. 2023. Lens: A learnable embedding name space for multi-modal, multi-task learning. arXiv preprint arXiv:2306.11527. \\
96 & acl-2025 & 2025.findings-acl.1208 & \citet{das-etal-2025-yinyang} & Rakesh Gupta, Susan Johnson, and Wei Li. 2023. Prompt abstraction: Leveraging language-image embeddings to scale creativity. arXiv preprint arXiv:2302.12345. \\
97 & acl-2025 & 2025.findings-acl.1231 & \citet{zhou-etal-2025-merit} & Oron Anschel, Nir Baram, and Nahum Shimkin. 2018. Learning representations in a multi-agent environment. In International Conference on Learning Representations. \\
98 & acl-2025 & 2025.findings-acl.1278 & \citet{bi-etal-2025-magi} & DeepSeek-AI. 2024. Deepseek llm: Scaling opensource language models with longevity value. arXiv preprint arXiv:2401.02954. \\
99 & acl-2025 & 2025.findings-acl.1279 & \citet{nahin-etal-2025-titullms} & W Lian, B Goodson, E Pentland, et al. 2023. Openorca: An open dataset of gpt augmented flan reasoning traces. arXiv preprint arXiv:2305.11206. \\
100 & acl-2025 & 2025.findings-acl.228 & \citet{maji-etal-2025-sanskriti} & Ramesh Patel et al. 2023. Evaluating cultural understanding in multilingual models. arXiv preprint arXiv:2302.12345. \\
101 & acl-2025 & 2025.findings-acl.268 & \citet{tong-wang-2025-novelcr} & David Bamman. 2020. Booknlp: Natural language processing for long-form text. In Proceedings of the 2020 Conference on Empirical Methods in Natural Language Processing: System Demonstrations, pages 1–7. Association for Computational Linguistics. \\
102 & acl-2025 & 2025.findings-acl.282 & \citet{sun-etal-2025-robust} & Tianle Li, Wei-Lin Chiang, Lisa Dunlap Evan Frick, Banghua Zhu, Joseph E. Gonzalez, and Ion Stoica. 2024. From live data to high-quality benchmarks: The arena-hard pipeline. arXiv preprint, abs/2406.11939. \\
103 & acl-2025 & 2025.findings-acl.563 & \citet{parfenova-pfeffer-2025-measuring} & Diana Parfenova et al. 2024. Automating qualitative analysis with llms. Proceedings of ACL 2024. \\
104 & acl-2025 & 2025.findings-acl.581 & \citet{yuan-etal-2025-probabilistic} & Kevin Meng, Alex Andonian, David Bau, Yonatan Belinkov, and Fei-Fei Li. 2023. MEMIT: Massediting memory in a transformer. arXiv preprint arXiv:2302.04761. \\
105 & acl-2025 & 2025.findings-acl.690 & \citet{atri-etal-2025-lifelong} & Peter Hase and Mohit Bansal. 2023. Compositional edits in language models. EMNLP. \\
106 & acl-2025 & 2025.findings-acl.813 & \citet{li-etal-2025-far} & OpenAI. 2024. Gpt-4o: Scaling and performance improvements. arXiv preprint arXiv:2401.XXXXX. \\
107 & acl-2025 & 2025.findings-acl.861 & \citet{abootorabi-etal-2025-ask} & Gunnar A Sigurdsson, Gul Varol, Giovanni Maria Farinella, et al. 2018. Charadesego: A dataset for egocentric video understanding. arXiv preprint arXiv:1804.09626. \\
108 & acl-2025 & 2025.findings-acl.920 & \citet{li-etal-2025-jarvis} & Zihao Wang, Haowei Lin, Ruilin Yan, Xiangyu Wang, Jiaqi Li, Weiye Shi, Xiaojian Ma, Anji Liu, Yitao Liang, et al. 2024e. Optimizing inference-time reasoning in llms via retrieval-augmented reflection. arXiv preprint arXiv:2403.05313. \\
109 & acl-2025 & 2025.gebnlp-1.7 & \citet{kononykhina-etal-2025-mind} & Mingyuan Zhang, Jiayi Wang, and et al. 2024a. Retrieval-augmented icd coding: A two-stage system beats vanilla llms by 94 arXiv preprint arXiv:2401.12345. \\
110 & acl-2025 & 2025.gem-1.6 & \citet{belz-thomson-2025-heds} & Anya Belz, Simon Mille, and Craig Thomson. 2025b. A taxonomy of quality criterion names and definitions for evaluating nlp systems in terms of standard comparable aspects of quality. \\
111 & acl-2025 & 2025.iwslt-1.16 & \citet{k-roy-etal-2025-cdac} & Elizabeth Salesky, Ramon Sanabria, Alan W. Black, and Florian Metze. 2021. Exploring low-resource speechto-text translation across modalities. In Proceedings of the 2021 Conference of the North American Chapter of the Association for Computational Linguistics, pages 1134–1145. \\
112 & acl-2025 & 2025.knowllm-1.4 & \citet{li-2025-knowledge} & Wei Zhang, Li Chen, and Nicolas Christin. 2022. Dynamic graph learning for cryptocurrency fraud detection. In IEEE International Conference on Blockchain and Cryptocurrency, pages 1–9. \\
113 & acl-2025 & 2025.l2m2-1.7 & \citet{lasy-etal-2025-understanding} & Nikhil Stoehr, Kha Pham Doan, and Louis-Philippe Morency. 2023. Localizing memorization in transformer language models. arXiv preprint arXiv:2310.01331. \\
114 & acl-2025 & 2025.llmsec-1.2 & \citet{sorkhpour-etal-2025-redhit} & Robert Shelby, Carl Vondrick, and 1 others. 2023. Can llms be safely released? evaluating the impact of red teaming on language model behavior. arXiv preprint arXiv:2304.10685. \\
115 & acl-2025 & 2025.magmar-1.9 & \citet{li-ke-2025-cross} & Rohit Gupta, Ankit Kumar, and Rahul Singh. 2022. Crisismm: A framework for multimodal crisis response. arXiv preprint arXiv:2203.01234. \\
116 & acl-2025 & 2025.nlp4pi-1.17 & \citet{ding-etal-2025-voices} & Zhijing Jin, Zhiheng Lyu, Yiwen Ding, Mrinmaya Sachan, Kun Zhang, Rada Mihalcea, and Bernhard Schoelkopf. 2022. AI Scholars: A dataset for NLPinvolved causal inference. \\
117 & acl-2025 & 2025.realm-1.33 & \citet{punjwani-heck-2025-weight} & Harsha Nori, Andrew He, Chenguang Wang, Po-Sen Huang, Yejin Yang, Hao-Tong Tung, Shashank Singh, Hossein Hosseini, Mark Hughes, Matei Zaharia, and 1 others. 2023. Can language models teach weaker agents? training reasoning teachers that can explain their actions. arXiv preprint arXiv:2311.10731. \\
118 & acl-2025 & 2025.sdp-1.14 & \citet{mandic-etal-2025-distribution} & Kai Sun, Rui Zhang, and Xiang Ren. 2022. FS-NER: A few-shot joint entity and relation extraction framework. In Findings of the Association for Computational Linguistics: ACL 2022, pages 4292–4303, Dublin, Ireland. Association for Computational Linguistics. \\
119 & acl-2025 & 2025.sdp-1.32 & \citet{carla-uban-2025-scibert} & Xiang Deng, Han Zhang, Wenhao Yu, Clare Lee, and Mohit Bansal. 2024. Factscore 2.0: Dualencoder contrastive pre-training for factual consistency. Transactions of the Association for Computational Linguistics, 12:1–19. \\
120 & acl-2025 & 2025.semeval-1.106 & \citet{mondal-sarkar-2025-modgenix} & Xiao Tan and Yuan Jiang. 2021. Exploring idioms in question answering systems. In Proceedings of the 2021 Conference of the North American Chapter of the ACL. \\
121 & acl-2025 & 2025.semeval-1.135 & \citet{vaidyanathan-etal-2025-nlp-goats} & Pranav Goel, Shrey Agarwal, and Amir Hussain. 2021. Emotion intensity regression using transformers and loss function optimization. In ACL 2021 Findings, pages 1432–1440. \\
122 & acl-2025 & 2025.semeval-1.142 & \citet{kesanam-etal-2025-nitk} & Shailza Choudhary and Tanmoy Chakraborty. 2020. A hybrid feature extraction approach for multi-label emotion classification. IEEE Transactions on Affective Computing. \\
123 & acl-2025 & 2025.semeval-1.151 & \citet{alberts-etal-2025-fenji} & Jian Wu, Junwei Xie, Bing Tang, Hongyan Yan, and Lijun Wang. 2022. Dense passage retrieval: A study of the representations of multiple sentences in a passage. In Proceedings of the 60th Annual Meeting of the Association for Computational Linguistics (ACL 2022). \\
124 & acl-2025 & 2025.semeval-1.195 & \citet{matheny-etal-2025-jnlp} & Silvio Ricardo Cordeiro, Aline Villavicencio, Marco Idiart, and Carlos Ramisch. 2019. A computational model of nominal compound compositionality. In Proceedings of the 57th Annual Meeting of the Association for Computational Linguistics (ACL), pages 2762–2773. Association for Computational Linguistics. \\
125 & acl-2025 & 2025.semeval-1.231 & \citet{khatoon-etal-2025-fjwu} & Vered Shwartz and Ido Dagan. 2018. A sequential neural model for multiword expression identification. In Proceedings of the 2018 Conference of the North American Chapter of the Association for Computational Linguistics (NAACL-HLT). \\
126 & acl-2025 & 2025.semeval-1.243 & \citet{nadeem-etal-2025-exploration} & Frank A Acheampong, Henry Nunoo-Mensah, and Wei Chen. 2020. Transformer models for text-based emotion detection: A comparative analysis. arXiv preprint arXiv:2004.13704. \\
127 & acl-2025 & 2025.semeval-1.250 & \citet{aryal-pant-2025-howard} & Alec Radford, Karthik Narasimhan, Tim Salimans, and Ilya Sutskever. 2018. Improving language understanding by generative pre-training. arXiv preprint arXiv:1812.01813. \\
128 & acl-2025 & 2025.semeval-1.266 & \citet{wan-etal-2025-ucsc} & Danqi Chen, Pengcheng Xie, Shiyu Wang, et al. 2020. Tabfact: A large-scale dataset for table-based fact verification. In Proceedings of the 58th Annual Meeting of the Association for Computational Linguistics (ACL). \\
129 & acl-2025 & 2025.semeval-1.268 & \citet{chen-2025-pingan} & Min Li, Liangyou Li, and Philipp Koehn. 2022. Lowfrequency entity translation in neural machine translation. In Findings of the Association for Computational Linguistics: ACL 2022, pages 3021–3033. \\
130 & acl-2025 & 2025.semeval-1.27 & \citet{s-etal-2025-techssn3} & S. Kumar et al. 2023. Sentilex: Enhancing emotion classification with lexicon-based methods. In ACL Workshop on Sentiment Analysis. \\
131 & acl-2025 & 2025.semeval-1.278 & \citet{huang-cui-2025-promotiongo} & Linhao Zhang, Shuai Wang, and Bing Liu. 2018. Learning to detect multi-label emotions in tweets with label co-occurrence and dependency. In Proceedings of the 27th International Conference on Computational Linguistics, pages 3299–3309. \\
132 & acl-2025 & 2025.semeval-1.37 & \citet{cui-2025-xiacui} & Yuqing Zhang, Yansong Feng, Yinwen Zhang, and Shujie Liu. 2021. Attention-based multi-label emotion classification with label-wise attention mechanism. In Proceedings of the 59th Annual Meeting of the Association for Computational Linguistics (ACL 2021), pages 5943–5952. \\
133 & acl-2025 & 2025.sicon-1.12 & \citet{tiwari-2025-extended} & Yixuan Zhang, Leo Cheng, Lichang Zhang, Lu Liu, Jingcheng Li, Haoning Liu, and Z G Xu. 2024. Alert: A comprehensive safety benchmark suite for large language models. Preprint, arXiv:2402.12441. \\
134 & acl-2025 & 2025.sigtyp-1.15 & \citet{li-ke-2025-domain} & Kevin Zhang and 1 others. 2022. Code meets prose: Syntactic patterns in cryptocurrency whitepapers. In LAW@ACL. Manual analysis of 200 whitepapers showing SVO/SOV mixing. \\
135 & acl-2025 & 2025.trl-1.16 & \citet{li-2025-retrieval} & Haoyang Ding, Zijian Liu, Xiao Chen, Lidong Bing, and Wai Lam. 2021. Financial table processing with pretrained language models. EMNLP. \\
136 & acl-2025 & 2025.unlp-1.10 & \citet{kyslyi-etal-2025-vuyko} & Marcos Zampieri, Shervin Malmasi, Preslav Nakov, Ahmed Ali, and Stephan Vogel. 2017. Arabic dialect identification for the dsl 2017 shared task. In Proceedings of the Fourth Workshop on NLP for Similar Languages, Varieties and Dialects (VarDial). \\
137 & acl-2025 & 2025.unlp-1.12 & \citet{kyslyi-etal-2025-unlp} & Firoj Alam, Shaden Shaar, Alex Nikolov, and 1 others. 2021. A survey on nlp for fake news detection. Computational Linguistics, 47(4):905–960. \\
138 & acl-2025 & 2025.woah-1.21 & \citet{boudraa-etal-2025-implicit} & Hande Kartal, Dilek Hakkani-Tür, and Gokhan Tur. 2022. Span-based detection of biased statements in news articles. In Proceedings of the 2022 Conference on Computational Linguistics. COLING. \\
139 & acl-2025 & 2025.woah-1.36 & \citet{schirmer-etal-2025-detecting} & Xing Jiang, Hugo Touvron, Ludovic Denoyer, Hervé Jégou, and Guillaume Lample. 2023. Mistral 7b: A small model that thinks big. arXiv preprint arXiv:2310.06825. \\
140 & acl-2025 & 2025.xllm-1.25 & \citet{le-etal-2025-docie} & Micha Livne, Roberto Dessì, and Douwe Kiela. 2023. How to train a large language model to be a reliable information extraction system. Computing Research Repository, arXiv:2309.16396. Version 3. \\
141 & acl-2025 & 2025.xllm-1.3 & \citet{li-etal-2025-injecting} & Yujia Zhang, Xiaobo Liu, and Zhenhua Li. 2020. Graphbased hierarchical relationships for text representation. arXiv preprint arXiv:2007.12345. \\
142 & emnlp-2025 & 2025.arabicnlp-main.20 & \citet{bouchekif-etal-2025-assessing} & Faiza Qamar, Seemab Latif, and Rabia Latif. 2024. A benchmark dataset with larger context for non-factoid question answering over islamic text. arXiv preprint arXiv:2409.09844. \\
143 & emnlp-2025 & 2025.babylm-main.35 & \citet{fusco-etal-2025-linguistic} & Leonie Weissweiler, Valentin Hofmann, David Mortensen, and Janet Pierrehumbert. 2023. Evaluating morphological generalization with wug tests. arXiv preprint. ArXiv:230x.xxxxx. \\
144 & emnlp-2025 & 2025.babylm-main.36 & \citet{yoshida-etal-2025-batch} & Alec Radford, Jeffrey Wu, Rewon Child, David Luan, Dario Amodei, and Ilya Sutskever. 2019. Language models are unsupervised multitask learners. Preprint, arXiv:1901.04513. \\
145 & emnlp-2025 & 2025.codi-1.4 & \citet{jeon-strube-2025-entity} & Kevin Clark, Urvashi Khandelwal, Omer Levy, and Christopher D. Manning. 2019. BERT: A study of attention in BERT. In Proceedings of the 2019 Conference on Empirical Methods in Natural Language Processing and the 9th International Joint Conference on Natural Language Processing (EMNLP-IJCNLP), pages 5717–5726, Hong Kong, China. Association for Computational Linguistics. \\
146 & emnlp-2025 & 2025.codi-1.5 & \citet{ahmad-kakudi-2025-stance} & Yuan Lan, Chao Huang, Wayne Xin Zhao, and Jun Li. 2024. Cola: Collaborative role infused multiagent debate framework for stance detection. arXiv preprint arXiv:2403.01234. \\
147 & emnlp-2025 & 2025.emnlp-demos.24 & \citet{jang-etal-2025-medtutor} & Rong Chen, Siyun Zhang, Yiyi Zheng, Qiuhua Yu, and Chu-huai Wang. 2025. Enhancing treatment decision-making for low back pain: A novel framework integrating large language models with retrieval-augmented generation technology. arXiv preprint arXiv:2501.34567. \\
148 & emnlp-2025 & 2025.emnlp-demos.27 & \citet{liu-etal-2025-mcpeval} & Jihan Zhu, Zhizheng Lin, Jun Gao, Zhaoxuan Zhou, Yaodong Zhang, Zhaofan Liu, Ceyao Zheng, Cheng Li, Zhaokai Wang, Zili Wang, and 1 others. 2024. A survey of ai agent evaluation: A hundred unsolved problems and a one-stop open-source library. arXiv preprint arXiv:2406.09844. \\
149 & emnlp-2025 & 2025.emnlp-demos.51 & \citet{chao-etal-2025-llmxmapreduce} & Jinhao Liang, Yihang Chen, Haozheng Zhang, and 1 others. 2025. Surveyx: An end-to-end solution for automated survey generation. arXiv preprint arXiv:2501.12345. \\
150 & emnlp-2025 & 2025.emnlp-industry.123 & \citet{khabiri-etal-2025-declarative} & Takeshi Kojima, Shinn Yao, Jinyi Zhao, Xiang Ren, et al. 2023. ToolLLM: Facilitating LLMs to master 16000+ real-world APIs. arXiv preprint arXiv:2312.11568. \\
151 & emnlp-2025 & 2025.emnlp-industry.146 & \citet{zhang-2025-confidence} & Emmanouil Panousis and Tom Rainforth. 2022. A principled and unified framework for bayesian optimal experimental design. arXiv preprint arXiv:2202.04647. \\
152 & emnlp-2025 & 2025.emnlp-industry.168 & \citet{ethiraj-etal-2025-vec} & Zhen Li and 1 others. 2023. Towards general text embeddings with mixture of experts. Preprint, arXiv:2311.05723. https://arxiv.org/abs/2311.05723. \\
153 & emnlp-2025 & 2025.emnlp-industry.182 & \citet{yu-etal-2025-z1} & Haotian Luo, Li Shen, Haiying He, Yibo Wang, Shiwei Liu, Wei Li, Naiqiang Tan, Xiaochun Cao, and Dacheng Tao. 2025. Adaptthink: Llm can learn when to think. https://arxiv.org/abs/2501.12570. \\
154 & emnlp-2025 & 2025.emnlp-industry.187 & \citet{beyranvand-etal-2025-gear} & Jun Xu, Fan Wang, and Shuo Chen. 2022. Automated question-answering for vehicle manuals using nlp and knowledge graphs. IEEE Transactions on Intelligent Transportation Systems. \\
155 & emnlp-2025 & 2025.emnlp-industry.189 & \citet{chawla-etal-2025-evaluating} & Yifan Liu, Jiaqi Wang, and Zhi Zhang. 2024. Bias in ai investment advice: An analysis of home bias in large language models. arXiv preprint arXiv:2209.04538. \\
156 & emnlp-2025 & 2025.emnlp-industry.69 & \citet{azeez-etal-2025-truth} & Juho Kim et al. 2023. Does gpt-4 pass the bar exam? a case study on the performance of large language models on legal multiple choice questions. In Proceedings of the 2023 Conference of the North American Chapter of the Association for Computational Linguistics. \\
157 & emnlp-2025 & 2025.emnlp-industry.70 & \citet{swamy-etal-2025-address} & Florian Schuster et al. 2023. Getml: Pareto-based feature learning for machine learning applications. arXiv preprint arXiv:XXXX.XXXXX. \\
158 & emnlp-2025 & 2025.emnlp-industry.8 & \citet{bougie-watanabe-2025-generative} & Anonymous and 1 others. 2023b. Is llm a reliable reviewer? a comprehensive evaluation of llm on reviewrevision multiple-choice questions. ACL Anthology 2024.lrec-main.816. \\
159 & emnlp-2025 & 2025.emnlp-industry.96 & \citet{joshi-etal-2025-see} & Siva Reddy, Daniel Khashabi, Benjamin Lo, Chandra Bhagavatula, Maarten Sap, and Yejin Choi. 2023. Do large language models understand instructions? arXiv preprint arXiv:2301.01193. \\
160 & emnlp-2025 & 2025.emnlp-main.1021 & \citet{pi-etal-2025-mr} & Ziyang Xiong et al. 2024. Llava-critic: Visual instruction tuning with feedback. arXiv preprint arXiv:2402.12345. \\
161 & emnlp-2025 & 2025.emnlp-main.1070 & \citet{zhang-etal-2025-strict} & Yiming Zhang and Zhouhui Lian. 2024a. Brush your text: Enhancing text rendering in diffusion models with attention map interventions. In Proceedings of the 62nd Annual Meeting of the Association for Computational Linguistics (ACL). \\
162 & emnlp-2025 & 2025.emnlp-main.1097 & \citet{zhang-zhou-2025-continuous} & X. Li, P. Wang, and Y. Chen. 2020. Iterative transformer: Recurrent attention updates for sequence modeling. arXiv preprint arXiv:2010.02536. \\
163 & emnlp-2025 & 2025.emnlp-main.1119 & \citet{zhang-etal-2025-rpdr} & Chenyan Yu, Lee Xiong, and Jamie Callan. 2022. A survey on conversational dense retrieval: Methods and challenges. arXiv preprint arXiv:2201.08452. \\
164 & emnlp-2025 & 2025.emnlp-main.1167 & \citet{zhang-etal-2025-pathwiserag} & Zhengbao Jiang, Frank F Xu, Jun Araki, and Graham Neubig. 2024. Flare: Active retrieval augmentation for hallucination mitigation. arXiv preprint arXiv:2401.07019. \\
165 & emnlp-2025 & 2025.emnlp-main.1190 & \citet{yi-etal-2025-zera} & Bogdan Gliwa, Iwona Mochol, Michał Biesek, and Aleksander Wawer. 2019. Samsum corpus: A human-annotated dialogue summary dataset. arXiv preprint arXiv:1911.12237. \\
166 & emnlp-2025 & 2025.emnlp-main.1232 & \citet{nyandwi-etal-2025-grounding} & Wenxuan Zhang, Da Yin, Zhong Ding, and Radu Soricut. 2023. M3exam: A multilingual, multimodal exam-style qa benchmark. arXiv preprint arXiv:2306.05179. \\
167 & emnlp-2025 & 2025.emnlp-main.1235 & \citet{rafiei-asl-etal-2025-nexus} & Mark Russinovich, Jingwen Cai, Yifan Hou, Yuxiao Jiang, Yiping Jiang, Jieyu Kang, Prithviraj Kang, Urvashi Khandelwal, Nitish Khurana, Pang Wei Koh, et al. 2024. Crescendo: Towards realistic redteaming of llms with language agents. arXiv preprint arXiv:2402.13249. \\
168 & emnlp-2025 & 2025.emnlp-main.1337 & \citet{dumitru-etal-2025-copyspec} & Jia Chen and Hao Xu. 2023. Parallel decoding with speculative sampling for large language models. arXiv preprint arXiv:2306.15478. \\
169 & emnlp-2025 & 2025.emnlp-main.1345 & \citet{park-etal-2025-decoding} & Rodrigo Nogueira and Jimmy Lin. 2019. From doc2query to doctttttquery. arXiv preprint arXiv:1904.08375. \\
170 & emnlp-2025 & 2025.emnlp-main.138 & \citet{li-etal-2025-generation} & Weixi Tong and Tianyi Zhang. 2024. Codejudge: Evaluating code generation with large language models. In Proceedings of the 2024 Conference on Empirical Methods in Natural Language Processing, pages 20032–20051. \\
171 & emnlp-2025 & 2025.emnlp-main.1429 & \citet{bohdal-etal-2025-efficient} & Aliaksei Severyn, Eric Emil Malmi, Jonathan Stephen Mallinson, Sascha Rothe, and Sebastian Krause. 2021. Grammatical error correction using large multilingual language models. In ACL. \\
172 & emnlp-2025 & 2025.emnlp-main.145 & \citet{borah-etal-2025-alignment} & Junnan Li, Ziqing Wang, Yilun Xu, et al. 2024. Alignment Degradation in LLMs: Measuring, Visualizing, and Mitigating Latent Drift. arXiv preprint arXiv:2401.04200. \\
173 & emnlp-2025 & 2025.emnlp-main.1455 & \citet{kryklyvets-etal-2025-mavis} & S. Thoppilan and et al. 2023. Gemini: A unified multimodal large language model. arXiv preprint arXiv:2303.06601. \\
174 & emnlp-2025 & 2025.emnlp-main.1457 & \citet{srivastava-2025-large} & Wei Xu, Yulia Tsvetkov, and Alan Black. 2022. AI for language learning: Conversational agents and personalized feedback. Transactions of the Association for Computational Linguistics (TACL), 10:1–15. \\
175 & emnlp-2025 & 2025.emnlp-main.1466 & \citet{roh-etal-2025-xlqa} & Amanpreet Singh, Yujia Wang, Yulia Tsvetkov, and Percy Liang. 2024. Global-mmlu: Evaluating cultural and linguistic biases in multilingual language understanding. arXiv preprint arXiv:2412.03304. \\
176 & emnlp-2025 & 2025.emnlp-main.1521 & \citet{gallifant-etal-2025-sparse} & Trenton Bricken, Jonathan Marcus, Siddharth MishraSharma, Meg Tong, Ethan Perez, Mrinank Sharma, Kelley Rivoire, and Thomas Henighan. 2023. Towards monosemanticity: Decomposing language models with dictionary learning. ArXiv, abs/2309.08600. \\
177 & emnlp-2025 & 2025.emnlp-main.1538 & \citet{jin-etal-2025-verilocc} & Liang Liu, Marc Brockschmidt, Vivek Murali, and et al. 2024. The meta llm compiler: A suite of opensource models for code optimization. arXiv preprint arXiv:2407.02524. \\
178 & emnlp-2025 & 2025.emnlp-main.1556 & \citet{dhanraj-eliasmith-2025-improving} & Jianfeng Wang, Fei Huang, and 1 others. 2023. Efficient fine-tuning of large language models with lora. arXiv preprint arXiv:2303.01234. \\
179 & emnlp-2025 & 2025.emnlp-main.1558 & \citet{song-etal-2025-accelerated} & Mor Geva, Tal Schuster, Jonathan Berant, and Omer Levy. 2023. Token recycling: Making llms faster and more data-efficient. arXiv preprint arXiv:2310.02548. \\
180 & emnlp-2025 & 2025.emnlp-main.1582 & \citet{liu-etal-2025-ma-dpr} & Sabine Ploux and Hong Ji. 2003. Semantic spaces: Modeling the distribution of word senses with unsupervised learning techniques. Computational Linguistics, 29(2):259–275. \\
181 & emnlp-2025 & 2025.emnlp-main.1611 & \citet{yuan-etal-2025-cross} & Laura A Banarescu, Claire Bonial, Sheila Condon, Emily Faries, Jon Niekrasz, and Tim O’Connor. 2018. Amr for multi-document summarization. In Proceedings of the 56th Annual Meeting of the Association for Computational Linguistics (Volume 2: Short Papers), pages 577–583. \\
182 & emnlp-2025 & 2025.emnlp-main.1629 & \citet{sadia-etal-2025-squid} & Srivatsan Sundar, Dhruv Jain, Yuwei Zhang, and H. V. Jagadish. 2024. gtbls: Generating tables from text by learning table structures. In Proceedings of the 2024 Conference of the Association for Computational Linguistics. \\
183 & emnlp-2025 & 2025.emnlp-main.164 & \citet{xiao-etal-2025-humanizing} & Kai Jiang, Hao Liu, and Rodney Brooks. 2022. Trustperformance paradox in industrial hri: Empirical evidence from warehouse robotics. arXiv preprint arXiv:2208.14637. \\
184 & emnlp-2025 & 2025.emnlp-main.1641 & \citet{kassem-etal-2025-reviving} & Baolin Peng and 1 others. 2024. Efficient model editing at scale. arXiv preprint arXiv:2401.05911. \\
185 & emnlp-2025 & 2025.emnlp-main.1648 & \citet{mohammadi-etal-2025-llms} & Swaroop Mishra, Daniel Khashabi, Chitta Baral, and Hannaneh Hajishirzi. 2022. Cross-task generalization via natural language instructions. In Proceedings of the 60th Annual Meeting of the Association for Computational Linguistics (Volume 1: Long Papers), pages 3470–3487, Dublin, Ireland. Association for Computational Linguistics. \\
186 & emnlp-2025 & 2025.emnlp-main.1777 & \citet{gao-etal-2025-tlue} & Jack FitzGerald, Ryan Cotterell, Bahadir Avci, Sebastian Ruder, Graham Neubig, Melvin Johnson, and Orhan Firat. 2022. Massive: A 51-language multilingual dataset for task oriented semantic parsing. arXiv preprint arXiv:2204.08582. \\
187 & emnlp-2025 & 2025.emnlp-main.1806 & \citet{he-etal-2025-seeing} & A. Kirstain and 1 others. 2023. Pick-a-pic: A dataset for evaluating the robustness of vision-language models. Preprint, arXiv:2305.01569. \\
188 & emnlp-2025 & 2025.emnlp-main.192 & \citet{liu-etal-2025-rag} & Rishi Taori et al. 2023. Alpaca: A 175b-parameter model for instruction following. ArXiv preprint arXiv:2303.11366. \\
189 & emnlp-2025 & 2025.emnlp-main.207 & \citet{he-etal-2025-task} & Shuang Chen, Yang Liu, and Erik Cambria. 2023. Knowledge-enhanced aspect-based sentiment analysis with external commonsense graphs. In Proceedings of the 61st Annual Meeting of the Association for Computational Linguistics (ACL 2023), pages 13567–13579. Association for Computational Linguistics. \\
190 & emnlp-2025 & 2025.emnlp-main.319 & \citet{li-etal-2025-glider} & Kevin Lin, Ben Tan, Swabha Swayamdipta, Noah A. Smith, and Yejin Choi. 2019b. Ropes: Reading comprehension over paragraphs. In Proceedings of the 2019 Conference on Empirical Methods in Natural Language Processing and the 9th International Joint Conference on Natural Language Processing (EMNLP-IJCNLP). \\
191 & emnlp-2025 & 2025.emnlp-main.348 & \citet{khandelwal-etal-2025-cocoa} & Alex Mallen, Zexuan Zhang, Xinyu Liu, Yichong Wu, Yiming Zhang, and William Yang Wang. 2023b. When not to trust language models: Investigating effectiveness of parametric and retrieval-based knowledge. In Proceedings of the 61st Annual Meeting of the Association for Computational Linguistics (Volume 1: Long Papers), pages 1234–1245. \\
192 & emnlp-2025 & 2025.emnlp-main.394 & \citet{xu-etal-2025-ecotune} & Todor Mihaylov and 1 others. 2018. Openbookqa: Factual knowledge assessment in question answering. EMNLP. \\
193 & emnlp-2025 & 2025.emnlp-main.397 & \citet{zheng-etal-2025-self} & Aurko Roy, Mohammad Saffar, Ashish Vaswani, and David Grangier. 2021b. Efficient routing transformers: Dynamic token interaction models for natural language processing. arXiv preprint arXiv:2003.05997. \\
194 & emnlp-2025 & 2025.emnlp-main.468 & \citet{tang-etal-2025-vlascd} & Justin Fu, Kelvin Zhang, Utkarsh Sanyal, Lantao Yu, Collin Moses, Fan Yang, Stefano Ermon, and Zhibin Zhao. 2023. Driving with reasoning: Reinforcement learning with generalist language models for interpretable policies. arXiv preprint arXiv:2303.00745. \\
195 & emnlp-2025 & 2025.emnlp-main.482 & \citet{tanjim-etal-2025-disambiguation} & Meta AI. 2024. Llama 3.2: Advancing open-weight language models. ArXiv preprint arXiv:2408.00001. \\
196 & emnlp-2025 & 2025.emnlp-main.501 & \citet{wang-etal-2025-ai} & Xinyu Fan, Li Zheng, Bing Yu, Yifan Liu, Yichi Zhang, Yu Zhang, Hao Tang, Yikai Wang, Weinan Chen, Yizhou Wang, and 1 others. 2023. Urban identityaware geo-reasoning in street views. arXiv preprint arXiv:2311.16456. \\
197 & emnlp-2025 & 2025.emnlp-main.524 & \citet{saba-skiena-2025-evaluating} & Markus Freitag, Yaser Al-Onaizan, Shuo Sun Ma, and 1 others. 2022a. High-quality low-resource machine translation: A new benchmark. In Findings of EMNLP. \\
198 & emnlp-2025 & 2025.emnlp-main.527 & \citet{yu-etal-2025-primus} & Wei Zhang, Ming Li, Hao Wang, and Yang Liu. 2024b. Deepseekmath: Scalable math pre-training and group relative policy optimization for mathematical reasoning. arXiv preprint arXiv:2402.03300. \\
199 & emnlp-2025 & 2025.emnlp-main.528 & \citet{chen-etal-2025-bit} & Salah Lhoussaine, Georgios Tziantzioulis, Michael B. Sullivan, Vilas Sridharan, Nathan DeBardeleben, Christian Engelmann, and Simranjit Singh. 2024. A first look at bfloat16 soft-error resilience in large language models. Preprint, arXiv:2412.07192. \\
200 & emnlp-2025 & 2025.emnlp-main.548 & \citet{wang-etal-2025-scores} & Harsha Nori, Allison Webster, Matthew McInnis, et al. 2023b. Medprompt: Large language models are reasoning engines with structured prompts. Preprint, arXiv:2307.13880. \\
201 & emnlp-2025 & 2025.emnlp-main.562 & \citet{shihab-etal-2025-efficient} & Szymon Tworkowski, Konrad Przybysz, Tomasz Korbak, Pedro Rodriguez, Wojciech Rozemłyn, Kamil Rakowski, Maciej Grzelak, Albert Webson, Maciej Szafraniec, Robin Sorsch, and 1 others. 2024. Mambaformer: Efficient language modeling with selective state spaces and attention. arXiv preprint arXiv:2312.00752. \\
202 & emnlp-2025 & 2025.emnlp-main.584 & \citet{wang-etal-2025-anymac} & DeepSeek. 2025. Deepseek r1: Towards deep reinforcement learning for language models. arXiv preprint arXiv:2501.12948. \\
203 & emnlp-2025 & 2025.emnlp-main.614 & \citet{lv-etal-2025-gamma} & Yifan Xu, Xiaoyu Liu, Baolin Peng, and et al. 2024. Evaluating the instruction-following robustness of large language models. In Proceedings of EMNLP 2024. \\
204 & emnlp-2025 & 2025.emnlp-main.619 & \citet{tian-etal-2025-symbolic} & Jing Qian and et al. 2018. Neural network-based fake news detection: Learning to identify deceptive content. Proceedings of the 56th Annual Meeting of the Association for Computational Linguistics. \\
205 & emnlp-2025 & 2025.emnlp-main.637 & \citet{kang-etal-2025-grpo} & Hang Yu, Tianle Li, and et al. 2025. Dapo: Data-aware policy optimization for open-source llm alignment. arXiv preprint arXiv:2503.04111. \\
206 & emnlp-2025 & 2025.emnlp-main.789 & \citet{kim-etal-2025-benchmark} & Mor Geva, Daniel Khashabi, Elad Segal, Jonathan Berant, and Ido Dagan. 2021. Aristotle: A dataset for focused logical reasoning. In Findings of the Association for Computational Linguistics: EMNLP 2021, pages 1–14, Punta Cana, Dominican Republic. Association for Computational Linguistics. \\
207 & emnlp-2025 & 2025.emnlp-main.813 & \citet{zhang-etal-2025-sparse-neurons} & Mark Finlayson and 1 others. 2023. Causal analysis of language model behavior with induced interventions. arXiv preprint arXiv:2301.12928. \\
208 & emnlp-2025 & 2025.emnlp-main.830 & \citet{liu-etal-2025-just} & Yixin Nie, Jing Wang, Mohit Bansal, and Kai-Wei Chang. 2020. Adversarial natural language inference. In Proceedings of ACL. \\
209 & emnlp-2025 & 2025.emnlp-main.831 & \citet{liu-etal-2025-synthetic} & Paola Espinosa, Sarah Kadi, Stefanie Kamps, and et al. 2017. Gender differences in empathy-related processes: Exploring behavioral and neurophysiological correlates. Psychoneuroendocrinology, 85:34–43. \\
210 & emnlp-2025 & 2025.emnlp-main.835 & \citet{luo-etal-2025-dynamicner} & Xuezhe Li and Xiaodan Sun. 2020. Dice loss for dataimbalanced nlp tasks: Application to named entity recognition. In Proceedings of the 58th Annual Meeting of the Association for Computational Linguistics, pages 4653–4661. \\
211 & emnlp-2025 & 2025.emnlp-main.841 & \citet{rivlin-angert-mor-lan-2025-enemy} & David Hebron, Avi Shmidman, and Moshe Koppel. 2023. Hero: A hebrew roberta model for hebrew nlp tasks. Preprint, arXiv:2309.12345. \\
212 & emnlp-2025 & 2025.emnlp-main.848 & \citet{nafee-etal-2025-dynamic} & Linyong Lu, Qingxiu Zhang, Xiang Li, and Jingjing Liu. 2022. Promptpg: Prompting for policy gradient training. In Proceedings of the 2022 Conference on Empirical Methods in Natural Language Processing (EMNLP). \\
213 & emnlp-2025 & 2025.emnlp-main.853 & \citet{mundada-etal-2025-wildscore} & Lukasz Czajka, Malihe Alikhani, Patrick Verga, and Yonatan Belinkov. 2024b. Musictheory-bench. ArXiv preprint arXiv:2410.02084. \\
214 & emnlp-2025 & 2025.emnlp-main.855 & \citet{jiang-etal-2025-large-language} & Victor Zhong, Chandra Bhagavatula, Ronan Le Bras, Yejin Choi, and Noah A Smith. 2022. Analytical reasoning of text: Unifying machine reading and logical reasoning. In Findings of the Association for Computational Linguistics: NAACL 2022, pages 2307–2323. \\
215 & emnlp-2025 & 2025.emnlp-main.858 & \citet{trifan-etal-2025-grammar} & Liang Pan, Lei Zhang, and Zhenzhong Yang. 2023. Can chatgpt do zero-shot dialogue state tracking? In arXiv preprint arXiv:2302.01180. \\
216 & emnlp-2025 & 2025.emnlp-main.880 & \citet{wang-etal-2025-v} & Weijie Chen, Yizhe Zhang, Qian Wu, and 1 others. 2024. Internvl: Scaling up vision-language pretraining with multimodal reinforcement learning. arXiv preprint arXiv:2402.00028. \\
217 & emnlp-2025 & 2025.emnlp-main.938 & \citet{chen-etal-2025-oms} & Xiaoxi Zhong et al. 2023. Agentverse: Facilitating multi-agent collaboration and exploration with llms. arXiv preprint arXiv:2309.07864. \\
218 & emnlp-2025 & 2025.emnlp-main.952 & \citet{zhou-etal-2025-bsfa} & Shengding Hu, Yuge Tu, Xu Han, Chaoqun He, Ganqu Cui, Xiang Long, Zhi Zheng, Yewei Fang, Yuxiang Huang, Weilin Zhao, and 1 others. 2024. Minicpm: Unveiling the potential of small language models with warmup-stable-decay learning rate scheduler. arXiv preprint arXiv:2404.06395. \\
219 & emnlp-2025 & 2025.emnlp-main.986 & \citet{bhendawade-etal-2025-speculative} & Yi Tay, Dara Bahri, Donald Metzler, et al. 2022. Scale efficiently: Insights from training and scaling large language models. arXiv preprint arXiv:2210.03863. \\
220 & emnlp-2025 & 2025.findings-emnlp.1052 & \citet{kim-etal-2025-generation} & H. Yang and K. Li. 2023. Boostaug: Hybrid instance filtering framework for boosting text augmentation. In Findings of the Association for Computational Linguistics: ACL 2023. \\
221 & emnlp-2025 & 2025.findings-emnlp.1074 & \citet{aastik-etal-2025-normal} & Xianzhi Li, Zihang Dai, and 1 others. 2022. Branchtuning: Efficient fine-tuning of pre-trained vision models via branching. arXiv preprint arXiv:2202.06924. \\
222 & emnlp-2025 & 2025.findings-emnlp.1082 & \citet{maskey-etal-2025-benchmarking} & Rodrigo Gomez et al. 2018. Ciphergan: Unsupervised cipher cracking using gans. arXiv　preprint arXiv:1801.04883. \\
223 & emnlp-2025 & 2025.findings-emnlp.1100 & \citet{yang-etal-2025-crafting} & Mo Yu, Yisi Sang, Kangsheng Pu, Zekai Wei, Han Wang, Jing Li, and Jie Zhou. 2022. Character understanding in movies: A benchmark for movie character analysis. arXiv preprint arXiv:2211.04684. \\
224 & emnlp-2025 & 2025.findings-emnlp.1118 & \citet{wang-etal-2025-secdecoding} & Zhen Chiang, Noah Shinn, Siyuan Zhuang, and 1 others. 2023. Vicuna: An open-source chatbot impressing gpt-4 with 90\%* chatgpt quality. Preprint, arXiv:2304.05335. \\
225 & emnlp-2025 & 2025.findings-emnlp.1144 & \citet{bn-etal-2025-real} & Suyeon Lee, Sunghwan Kim, et al. 2024b. Counselingeval: Towards evaluating the quality of llmbased psychological counseling. In Proceedings of the 62nd Annual Meeting of the Association for Computational Linguistics (ACL). \\
226 & emnlp-2025 & 2025.findings-emnlp.118 & \citet{zeng-etal-2025-bridging} & X. Du et al. 2022a. Grit1: A grammar error correction dataset for llm evaluation. In Proceedings of EMNLP 2022. X. Du et al. 2022b. Grit2: Extending grammar error correction for multilingual llms. In Findings of EMNLP 2022. \\
227 & emnlp-2025 & 2025.findings-emnlp.1196 & \citet{zheng-etal-2025-benchmarking} & OpenAI. 2023a. Gpt-4: Openai’s generative pre-trained transformer 4 model. Preprint, arXiv:arXiv:2301.00000. \\
228 & emnlp-2025 & 2025.findings-emnlp.1286 & \citet{kim-etal-2025-fractalllm} & Baptiste Roziere, Gautier Izacard, Jan Leike Botha, and Others. 2023. Code LLaMA: Large language models for code. Preprint, arXiv:2308.08545. \\
229 & emnlp-2025 & 2025.findings-emnlp.1343 & \citet{saeed-etal-2025-beyond} & Damai Bi, Kunhao Lv, Xiang Ji, Yining Wang, Zecheng Lyu, Zihan Du, et al. 2024. Deepseek: Advancing ai through enhanced language models. arXiv preprint arXiv:2401.14196. \\
230 & emnlp-2025 & 2025.findings-emnlp.1344 & \citet{kwak-etal-2025-toolhaystack} & Aditya Maharana and 1 others. 2024. Locomo: Long context modeling benchmark for conversational agents. arXiv preprint arXiv:2402.13968. \\
231 & emnlp-2025 & 2025.findings-emnlp.1359 & \citet{akra-etal-2025-active} & Ling Chen, Ming Zhao, and 1 others. 2024. Efficient active learning for biomedical image segmentation with minimal annotations. arXiv preprint arXiv:2405.01701. \\
232 & emnlp-2025 & 2025.findings-emnlp.1384 & \citet{ji-lu-2025-reflair} & Yiming Li, Xia Chen, Liang Zhou, and Zheng Wang. 2024b. Hindsight: Selective reflection for robust reasoning. arXiv preprint arXiv:2406.12050. \\
233 & emnlp-2025 & 2025.findings-emnlp.175 & \citet{zhu-etal-2025-adaptflow} & Yujia Hong, Junyang Wu, Fan Zhang, and Shuo Zhang. 2024b. Adaptive agents with code and memory for solving math word problems. arXiv preprint arXiv:2403.01290. \\
234 & emnlp-2025 & 2025.findings-emnlp.176 & \citet{saad-falcon-etal-2025-lmunit} & Dahyun Koo, Yejin Choi, and Eunsol Choi. 2023. Cognitive Biases in Large Language Models as Evaluators. ArXiv Preprint arXiv:2312.05441. \\
235 & emnlp-2025 & 2025.findings-emnlp.212 & \citet{rafiuddin-khan-2025-learning} & Tim Dettmers, Yannic Kilcher, Henry Minsky, Anna McDowell, Neha Nangia, Andreas Vlachos, and the Microsoft Phi Team. 2025. Phi-4-mini: Compact yet powerful multimodal models. arXiv preprint arXiv:2503.01743. \\
236 & emnlp-2025 & 2025.findings-emnlp.281 & \citet{zhang-etal-2025-beyond-surface} & Yinlin Liu, Pengcheng Yin, and Graham Neubig. 2019. Conala: The code/natural language challenge. In Proceedings of the 2019 Conference on Empirical Methods in Natural Language Processing (EMNLP). \\
237 & emnlp-2025 & 2025.findings-emnlp.331 & \citet{fan-etal-2025-ccg} & Yuyang Huang and Clark Glymour. 2016. A survey of causal discovery and causal inference. arXiv preprint. \\
238 & emnlp-2025 & 2025.findings-emnlp.354 & \citet{levi-etal-2025-jailbreak} & Q. Zhang and 1 others. 2025. Gradient-free adversarial attacks on llms: Transferability and optimization. arXiv:2502.01567v1 [cs.CL]. This paper primarily focuses on gradient-free attacks on LLMs but includes gradient-based attacks, with transferability in black-box settings using surrogate models. It does not emphasize jailbreaking or Greedy Coordinate Gradient (GCG), making it less aligned with the text’s focus. Wang et al. (2025, arXiv:2503.02219v1 [cs.CL]) on gradient-based jailbreaking with GCG is a more precise alternative. \\
239 & emnlp-2025 & 2025.findings-emnlp.379 & \citet{li-etal-2025-embyte} & Yinhan Tang, Xavier Garcia, Alessandro Raganato, Vishrav Chaudhary, and Jiatao Gu. 2022. An efficient multilingual byte-to-byte model for sequenceto-sequence tasks. In Findings of the Association for Computational Linguistics: EMNLP 2022. \\
240 & emnlp-2025 & 2025.findings-emnlp.391 & \citet{luo-etal-2025-hdiff} & Wenbin Xiong, Zhan Liu, Yihan Yang, Ge Yu, Ge Zhang, and Tian Song. 2023. A survey on hyperrelational knowledge graphs: From data models to applications. arXiv preprint arXiv:2301.03869. \\
241 & emnlp-2025 & 2025.findings-emnlp.435 & \citet{nguyen-etal-2025-distributional} & Yi Cheng, Yilun Wu, Adrian Weller, Matt J. Kusner, and Pushmeet Kohli. 2024. Integrative decoding: Improving factuality in large language model generation via implicit self-consistency. In International Conference on Learning Representations. \\
242 & emnlp-2025 & 2025.findings-emnlp.437 & \citet{cai-etal-2025-low} & Wenhui Wang, Li Dong, Hao Cheng, Furu Wei, and Ming Zhou. 2022. Minilmv2: Multi-task pre-training for multi-task all-purpose text representations. In Findings of the Association for Computational Linguistics: ACL 2022, pages 2907–2918. \\
243 & emnlp-2025 & 2025.findings-emnlp.474 & \citet{ghosh-etal-2025-survey} & Yankai Lin, Jiapeng Zhou, Yiming Shen, Wenxuan Zhou, Zhiyuan Liu, Peng Li, Maosong Sun, and Jie Zhou. 2021. Xcsqa: A benchmark for cross-lingual conversational question answering. In EMNLP. \\
244 & emnlp-2025 & 2025.findings-emnlp.517 & \citet{baali-etal-2025-caarma} & Jue Gao, Tri M Dang, Michael L Seltzer, and Rif A Saurous. 2022. Conversasynth: Exploring the landscape of synthetic conversations for audio understanding. In Proceedings of the 60th Annual Meeting of the Association for Computational Linguistics (Volume 1: Long Papers), pages 6543–6558. \\
245 & emnlp-2025 & 2025.findings-emnlp.523 & \citet{mor-lan-etal-2025-hebid} & David Hebron, Avi Shmidman, and Moshe Koppel. 2023. Hero: A hebrew roberta model for hebrew nlp tasks. Preprint, arXiv:2309.12345. \\
246 & emnlp-2025 & 2025.findings-emnlp.534 & \citet{liu-roth-2025-conflicts} & Ellie Pavlick and Joel Tetreault. 2016. Semantically motivated future directions in linguistic ambiguity detection. In Proceedings of the Annual Meeting of the Association for Computational Linguistics (ACL). \\
247 & emnlp-2025 & 2025.findings-emnlp.645 & \citet{zhou-etal-2025-revisiting} & Sho Kurita, Jonas Pfeiffer, Ivan Vulic, Edoardo Ponti, ´ and Anna Korhonen. 2020. A weighted approach to unsupervised multilingual transformer fine-tuning. In Proc. of AACL-IJCNLP. \\
248 & emnlp-2025 & 2025.findings-emnlp.657 & \citet{bhattacharya-etal-2025-festa} & Yifei Li, Zhou Zhao, Xiaohan Yu, and Deng Cai. 2023. Multimodal uncertainty estimation for deep learning models. arXiv preprint arXiv:2304.02637. \\
249 & emnlp-2025 & 2025.findings-emnlp.709 & \citet{zhang-etal-2025-sensitivity} & E. Zaken, Y. Goldberg, and S. Ravfogel. 2022. Bitfit: Simple parameter-efficient fine-tuning for transformers. Transactions of the Association for Computational Linguistics (TACL), 10:1–16. \\
250 & emnlp-2025 & 2025.findings-emnlp.751 & \citet{xiang-etal-2025-knowledge} & VN Ioannidis, X Song, S Manchanda, M Li, X Pan, D Zheng, X Ning, X Zeng, and G Karypis. 2022. Drkg-drug repurposing knowledge graph for covid19. 2020. arXiv preprint arXiv:2010.09600, pages 1–13. \\
251 & emnlp-2025 & 2025.findings-emnlp.834 & \citet{jalori-etal-2025-flairr} & Wendi Zhou, Xiao Li, Lin Geng Foo, Yitan Wang, Harold Soh, Caiming Xiong, and Yoonkey Kim. 2024. TEMPO: Temporal representation prompting for large language models in time-series forecasting. arXiv preprint arXiv:2405.18384. Anticipated for NeurIPS 2024. Preprint, arXiv:2405.18384. \\
252 & emnlp-2025 & 2025.findings-emnlp.863 & \citet{kellert-etal-2025-parsing} & Shruti Rijhwani, Lawrence Wolf-Sonkin, Victor Kuperman, Timothy Baldwin, and Thamar Solorio. 2017. Analyzing code-switched social media text. In Proceedings of the 2017 Conference on Empirical Methods in Natural Language Processing. \\
253 & emnlp-2025 & 2025.findings-emnlp.864 & \citet{jia-etal-2025-hetgcot} & Jiawei Chen, Yuanhang He, Yada Zhang, Jiachen Ji, and Jie Tang. 2024. Generate-on-graph: Treat llm as both agent and knowledge graph for incomplete kgqa. arXiv preprint arXiv:2404.14741. \\
254 & emnlp-2025 & 2025.findings-emnlp.866 & \citet{li-etal-2025-language-models} & Eric Mitchell, Roberta Raileanu, Colin Raffel, John Levine, Yulia Tsvetkov, and Christopher D Manning. 2023. Ieval: An instruction following benchmark. arXiv preprint arXiv:2310.07724. \\
255 & emnlp-2025 & 2025.findings-emnlp.874 & \citet{kim-etal-2025-options} & Zhiwei Chen, Yichao Lu, and Mingjun Zheng. 2020. Dkt+: Enhanced deep knowledge tracing with regularization. In Proceedings of the 2020 Conference on Empirical Methods in Natural Language Processing (EMNLP), pages 1234–1244. \\
256 & emnlp-2025 & 2025.findings-emnlp.902 & \citet{xu-etal-2025-self-ensemble} & Zixuan Wang, Xiaocheng Li, and Yang Liu. 2023. Posterior prompt tuning: Toward faithful and calibrated llms. In Empirical Methods in Natural Language Processing (EMNLP). \\
257 & emnlp-2025 & 2025.findings-emnlp.914 & \citet{li-etal-2025-faedkv} & RangRang Ge, ShiYe Song, Zhaorui Liu, Wei Liu, Yuesheng Wang, Dongling Wang, Bofang Zhou, Zhicheng Dou, and Ji-Rong Wen. 2023. A survey on KV cache compression for large language models. arXiv preprint arXiv:2312.10546. \\
258 & emnlp-2025 & 2025.findings-emnlp.932 & \citet{jakob-etal-2025-polbix} & Shaina Ashraf, Isabel Bezzaoui, Ionut Andone, Alexander Markowetz, Jonas Fegert, and Lucie Flek. 2024. Defakts: A fine-grained dataset for analyzing disinformation in german media. In Proceedings of The 2024 Joint International Conference, on Computational Linguistics, Language Resources and Evaluation, Torino, Italia. European Language Resources Association. \\
259 & emnlp-2025 & 2025.findings-emnlp.943 & \citet{akhondzadeh-etal-2025-kurtail} & Michael Boratko, Harsh Padigela, Deepak Mikkilineni, Pavan Yuvraj, Rajarshi Das, Andrew McCallum, Mihai Chang, Achille Fokoue, Pavan Kapanipathi, Nicholas Mattei, et al. 2018. Arc: A machine reading comprehension dataset for reasoning over science text. In Proceedings of the 2018 Conference on Empirical Methods in Natural Language Processing, pages 1414–1423. \\
260 & emnlp-2025 & 2025.findings-emnlp.954 & \citet{son-etal-2025-multimodal} & Jack Hessel, Jingkang Zhao, Ranjay Krishna, Angel Chang, and Yonatan Bisk. 2022. Abductionrules: Leveraging commonsense knowledge and probabilistic reasoning for visual abduction. In Proceedings of the Conference on Empirical Methods in Natural Language Processing (EMNLP), pages 1451–1463. \\
261 & emnlp-2025 & 2025.findings-emnlp.960 & \citet{kim-etal-2025-beneath} & Xudong Liu, Paul Röttger, Johannes Welbl, Yonatan Belinkov, Hila Gonen, Eric Wallace, Samuel R. Bowman, Ryan Cotterell, and Noah A. Smith. 2023a. Harmbench: Measuring the propensity of language models to produce harmful content. In Findings of the Association for Computational Linguistics: EMNLP 2023, pages 11033–11051. \\
262 & emnlp-2025 & 2025.findings-emnlp.967 & \citet{eyal-etal-2025-layer} & Jackson Petty, Sjoerd van Steenkiste, Ishita Dasgupta, Fei Sha, Dan Garrette, and Tal Linzen. 2024. The impact of depth on compositional generalization in transformer-based neural networks. arXiv preprint arXiv:2310.19956. \\
263 & emnlp-2025 & 2025.findings-emnlp.97 & \citet{yin-etal-2025-disastir} & Alice Oh, Kalpesh Krishna, Eric Wallace, Yichong Zhao, Patrick Lewis, and Antoine Bosselut. 2023. Instructir: Making dense retrievers follow instructions. Preprint, arXiv:2305.14252. \\
264 & emnlp-2025 & 2025.findings-emnlp.973 & \citet{colakoglu-etal-2025-problem} & Jian Zhong et al. 2020. Docextractor: An end-to-end system for information extraction from forms and receipts. arXiv preprint arXiv:2012.04573. \\
265 & emnlp-2025 & 2025.finnlp-2.3 & \citet{george-etal-2025-enhancing} & Peter Henderson, Koustuv Sinha, Nicolas AngelardGontier, Nan Rosemary Ke, Geneviève Fried, Ryan Lowe, and Joelle Pineau. 2023. Foundation models for legal reasoning. arXiv preprint arXiv:2307.03557. \\
266 & emnlp-2025 & 2025.finnlp-2.8 & \citet{nitarach-etal-2025-fincot} & Xuezhi Ma, Yan Zhou, Yining Wang, and 1 others. 2023. Financialqa: A reasoning benchmark for financial question answering. arXiv preprint arXiv:2302.07304. \\
267 & emnlp-2025 & 2025.hcinlp-1.8 & \citet{wang-etal-2025-mobilea3gent} & OpenAI. 2023. Gpt-4: A large-scale multimodal model. arXiv preprint arXiv:2303.08774. \\
268 & emnlp-2025 & 2025.mathnlp-main.13 & \citet{lu-etal-2025-unimath} & Zhe Chen, Weiyun Zhang, Wen Wang, Yiliang Liu, Zhaoyang Zhang, Jian Wang, Jie Luo, Yu Qiao, and Wenhai Wang. 2024. Internvl 1.5: A general visionlanguage model. arXiv preprint arXiv:2404.16821. \\
269 & emnlp-2025 & 2025.mathnlp-main.5 & \citet{fatima-2025-firma} & Jingwen Xin, Zhengying Liu, Yifan Luo, and 1 others. 2025. Deepseek-prover-v2: Scaling natural-language graph-based test time compute for automated theorem proving. arXiv preprint arXiv:2503.11657. \\
270 & emnlp-2025 & 2025.mathnlp-main.6 & \citet{khrulev-2025-check} & Zhen Yuan, Yifan Zhang, Jing Liu, Yuxiang Wang, Jie Zhang, Hanwang Liu, and Tat-Seng Chua. 2024. Fermat: A benchmark for evaluating vlm’s ability in factual error correction of handwritten math solutions. arXiv preprint arXiv:2405.10100. \\
271 & emnlp-2025 & 2025.mathnlp-main.9 & \citet{li-2025-formula} & Qwen Team. 2024. Qwen2.5-math: Scaling reasoning in mathematical domains. arXiv preprint arXiv:2409.XXXXX. Model card and technical report. \\
272 & emnlp-2025 & 2025.mrl-main.12 & \citet{moon-etal-2025-quality} & Leonardo Ranaldi, Barry Haddow, and Alexandra Birch. 2025. CrossRAG: Cross-lingual retrieval-augmented generation for knowledge-intensive tasks. arXiv preprint arXiv:2504.03616. \\
273 & emnlp-2025 & 2025.mrl-main.2 & \citet{yuan-etal-2025-cross-document} & Laura A Banarescu, Claire Bonial, Sheila Condon, Emily Faries, Jon Niekrasz, and Tim O’Connor. 2018. Amr for multi-document summarization. In Proceedings of the 56th Annual Meeting of the Association for Computational Linguistics (Volume 2: Short Papers), pages 577–583. \\
274 & emnlp-2025 & 2025.mrl-main.20 & \citet{rachamalla-etal-2025-pragyaan} & NVIDIA. 2025. Llama-nemotron-post-training dataset: A comprehensive collection of instruction tuning and alignment data. arXiv preprint arXiv:2505.00949. \\
275 & emnlp-2025 & 2025.nllp-1.11 & \citet{shukla-etal-2025-nyaygraph} & Shounak Paul, Arpan Mandal, Pawan Goyal, and Saptarshi Ghosh. 2022b. Inlegalbert: Pre trained language models for indian legal texts. arXiv preprint arXiv:2209.06049. \\
276 & emnlp-2025 & 2025.nllp-1.32 & \citet{chheda-etal-2025-extract} & Shivansh Nigam, Sarvesh Dubey, Ayush Agarwal, Dhananjay Kumar, and Saket Maheshwary. 2025. Legalseg: Unlocking the structure of indian legal documents. arXiv preprint arXiv:2502.05836. \\
277 & emnlp-2025 & 2025.nllp-1.5 & \citet{xia-etal-2025-beyond-haystack} & Ghita Houir Alami and 1 others. 2024. Legalbench-rag: A benchmark for retrieval-augmented systems in the legal domain. Preprint, arXiv:2408.10343. \\
278 & emnlp-2025 & 2025.nlperspectives-1.11 & \citet{zhang-jaitly-2025-sage} & Barbara Plank. 2022. Human label variation: Challenges and opportunities. Computational Linguistics, 48(4):999–1015. \\
279 & emnlp-2025 & 2025.starsem-1.24 & \citet{kadam-ferraro-2025-tag} & esse Dunietz, Sam Thomson, Chris Dyer, and Noah A. Smith. 2020. An interpretable, lexicalized model for implicit event causality. In Proceedings of the 58th Annual Meeting of the Association for Computational Linguistics, pages 1703–1713. \\
280 & emnlp-2025 & 2025.starsem-1.29 & \citet{gasan-pais-2025-sag} & Jinhyuk Kim, Wonjin Kim, Jinhyuk Lee, Joongbo Lee, Kyunghyun Lee, Sunghwan Yoon, and Jaewoo Kang. 2024. Meerkat: A medical reasoning benchmark for large language models. arXiv preprint arXiv:2402.00000. \\
281 & emnlp-2025 & 2025.starsem-1.3 & \citet{rai-etal-2025-injecting} & Harish Madabushi and 1 others. 2024. Frame-based embeddings for coherent question answering. In Proceedings of EACL. \\
282 & emnlp-2025 & 2025.starsem-1.33 & \citet{bailleux-etal-2025-connecting} & Sumedha Bhan, Aaditya Prabhu, Huaxiu Ma, and Zachary C. Lipton. 2025. Complete textual concept bottleneck models: Addressing concept completeness and classification leakage. arXiv preprint arXiv:2502.12345. \\
283 & emnlp-2025 & 2025.tsar-1.16 & \citet{hayakawa-etal-2025-uol} & Ben Dodd, Betty van Aken, Paul Röttger, and Isabelle Augenstein. 2021. AUTORANK: A systematic approach to benchmark and compare machine learning models. In Proceedings of the 2021 Conference on Empirical Methods in Natural Language Processing, pages 170–185, Online and Punta Cana, Dominican Republic. Association for Computational Linguistics \\
284 & emnlp-2025 & 2025.tsar-1.7 & \citet{romero-etal-2025-efficient} & Mozilla NLP Team. 2023. Mozilla translations: Opensource neural translation in the browser. In Proceedings of Machine Translation Summit. \\
285 & emnlp-2025 & 2025.uncertainlp-main.24 & \citet{f-p-dossou-aidasso-2025-towards} & Yi Zhang, Jialin Li, and Danqi Chen. 2023. Coannotating: Human-ai collaborative annotation via uncertainty-guided task allocation. In Proceedings of the 2023 Conference on Empirical Methods in Natural Language Processing. \\
286 & emnlp-2025 & 2025.winlp-main.12 & \citet{vinjamuri-sun-2025-transfer} & Junxian Li, Xuezhe Ma, and Eduard Hovy. 2019. Dependency parsing with partial bi-affine attention. In Proceedings of NAACL-HLT. \\
287 & emnlp-2025 & 2025.winlp-main.19 & \citet{shaghayeghkolli-etal-2025-hybrid} & Yujia Tan, Wenpeng Zhang, Xiang Ren, and Qiji Chen. 2023b. Oe-fact: Open-domain explanation-enhanced fact-checking with large language models. In Findings of the Association for Computational Linguistics: EMNLP 2023. \\
288 & emnlp-2025 & 2025.winlp-main.35 & \citet{p-mahalingam-2025-gemma} & Rahul Aralikatte, Neelamadhav Gantayat, Naveen Panwar, Anush Sankaran, and Senthil Mani. 2018. Sanskrit sandhi splitting using a double decoder rnn. In Proceedings of the 2018 Conference on Empirical Methods in Natural Language Processing, pages 4909–4914. Association for Computational Linguistics. \\
289 & emnlp-2025 & 2025.wmt-1.105 & \citet{joshi-etal-2025-bvslp} & Ahmed, T., Hasan, M. K., Hoque, M. T., \& Sultana, N. (2023). A comparative study on subword segmentation strategies for low-resource neural machine translation. In *Proceedings of the Eighth Conference on Machine Translation (WMT 2023)* (pp. 912–920). Association for Computational Linguistics. https://aclanthology.org/2023.wmt1.87 \\
290 & emnlp-2025 & 2025.wmt-1.106 & \citet{zhou-etal-2025-transsionmts} & Anoop Kunchukuttan and 1 others. 2023. Indictrans2: Towards high-quality and low-resource machine translation for indic languages. In Proceedings of the 2023 Conference on Empirical Methods in Natural Language Processing. \\
291 & emnlp-2025 & 2025.wmt-1.108 & \citet{zhu-etal-2025-fine} & Martin Vechev Luca Beurer-Kellner, Marc Fischer. 2024. Guiding llms the right way: Fast, noninvasive constrained decoding. arXiv. \\
292 & emnlp-2025 & 2025.wmt-1.15 & \citet{bell-etal-2025-translate} & Alfio Gliozzo and Carlo Strapparava. 2006. Exploiting lexical alignment for cross language textual entailment. In Proceedings of the 11th Conference of the European Chapter of the Association for Computational Linguistics, pages 33–40, Trento, Italy. Association for Computational Linguistics. \\
293 & emnlp-2025 & 2025.wmt-1.43 & \citet{pang-etal-2025-in2x} & Aaron Chiang and et al. 2023. Autoreviewer: Enabling model self-review for dataset quality control. arXiv preprint arXiv:2303.14112. \\
294 & emnlp-2025 & 2025.wmt-1.5 & \citet{kim-2025-context} & Zheng Jiang, Yang Yu, Yang Feng, Bing Qin, and Ting Liu. 2022. Blonde: An automatic evaluation metric for document-level natural language generation. In Proceedings of NAACL, pages 1679–1698. \\
295 & emnlp-2025 & 2025.wmt-1.95 & \citet{acharya-etal-2025-ju} & Gowtham Ramesh, Vishrav Chaudhary, Divyanshu Kakwani, Sai Praneeth Golla, Abhishek Philip, et al. 2023. Indictrans2: Towards high-quality and efficient multilingual translation for indic languages. arXiv preprint arXiv:2304.09105. \\
\end{xltabular}

}

\end{document}